\documentclass[10pt,twocolumn,letterpaper]{article}

\usepackage{wacv}
\usepackage{times}
\usepackage{epsfig}
\usepackage{graphicx}
\usepackage{amsmath}
\usepackage{amssymb}
\usepackage{bm}
\usepackage{booktabs}
\usepackage{caption}
\usepackage[usenames,dvipsnames]{xcolor}
\usepackage{algorithm,algpseudocode}
\usepackage{xspace}
\usepackage{enumitem}
\usepackage[accsupp]{axessibility}  % Improves PDF readability for those with disabilities.

\def\ie{\emph{i.e}\onedot}

\def\etal{\emph{et al}\onedot}
\newcommand{\glitter}{GliTr\xspace}

\definecolor{pakistangreen}{rgb}{0.0, 0.4, 0.0}
\definecolor{figblue}{rgb}{0.0, 0.3, 0.6}
\definecolor{figorange}{rgb}{0.8, 0.4, 0.0}

\newenvironment{tight_itemize}{
\begin{itemize}
  \setlength{\topsep}{0pt}
  \setlength{\itemsep}{2pt}
  \setlength{\parskip}{0pt}
  \setlength{\parsep}{0pt}
}{\end{itemize}}

\def\wacvPaperID{****} % Enter the WACV Paper ID here

\wacvfinalcopy % *** Uncomment this line for the final submission

\ifwacvfinal
\usepackage[breaklinks=true,bookmarks=false]{hyperref}
\else
\usepackage[pagebackref=true,breaklinks=true,colorlinks,bookmarks=false]{hyperref}
\fi

\newcommand{\minisection}[1]{\vspace{2mm}\noindent{\textbf{#1}}}

\begin{document}

%%%%%%%%% TITLE
\title{GliTr: Glimpse Transformers with Spatiotemporal Consistency\\for Online Action Prediction}

\author{Samrudhdhi B Rangrej$^{1}$ \quad Kevin J Liang$^{2}$ \quad Tal Hassner$^{2}$ \quad James J Clark$^{1}$ \\
$^{1}$McGill University \quad $^{2}$Meta AI \\
{\tt\small samrudhdhi.rangrej@mail.mcgill.ca}}

\maketitle
\thispagestyle{empty}

%%%%%%%%% ABSTRACT
\begin{abstract}
Many online action prediction models observe complete frames to locate and attend to informative subregions in the frames called glimpses and recognize an ongoing action based on global and local information. However, in applications with constrained resources, an agent may not be able to observe the complete frame, yet must still locate useful glimpses to predict an incomplete action based on local information only. In this paper, we develop Glimpse Transformers (\glitter), which observe only narrow glimpses at all times, thus predicting an ongoing action and the following most informative glimpse location based on the partial spatiotemporal information collected so far. In the absence of a ground truth for the optimal glimpse locations for action recognition, we train \glitter using a novel spatiotemporal consistency objective: We require \glitter to attend to the glimpses with features similar to the corresponding complete frames (\ie spatial consistency) and the resultant class logits at time $t$ equivalent to the ones predicted using whole frames up to $t$ (\ie temporal consistency). Inclusion of our proposed consistency objective yields $\sim10\%$ higher accuracy on the Something-Something-v2 (SSv2) dataset than the baseline cross-entropy objective. Overall, despite observing only $\sim 33\%$ of the total area per frame, \glitter achieves 53.02\% and 93.91\% accuracy on the SSv2 and Jester datasets, respectively.
\end{abstract}

%%%%%%%%% BODY TEXT
\vspace{-5mm}
\section{Introduction}
\begin{figure}
    \centering
    \includegraphics[page=1,trim={0 0 0 0},clip,width=0.9\linewidth]{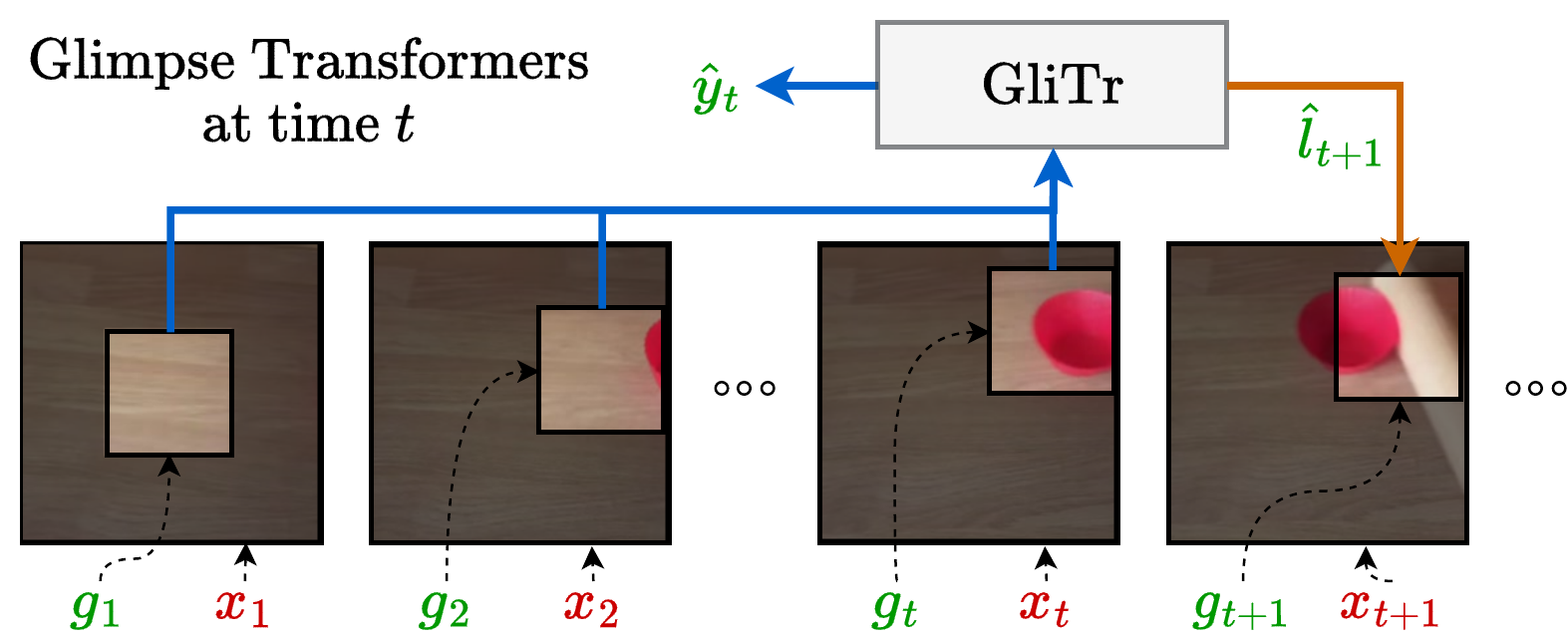}
    \caption{We propose \textbf{Glimpse Transformers (\glitter)}, an online action prediction model that only attends to the most informative glimpses ($g_t$) in the frames ($x_t$). While never observing frames completely, \glitter predicts label $\hat{y}_t$ (\ie an estimate of ongoing action at time $t$) and the next glimpse location $\hat{l}_{t+1}$ based solely on the glimpses observed up to $t$.}
    \vspace{-6mm}
    \label{fig:intro}
\end{figure}

Recent models such as TSM~\cite{lin2019tsm}, Swin-B~\cite{liu2022video}, or VideoMAE~\cite{tong2022videomae} have achieved impressive performance on video action recognition benchmarks, but they often make several assumptions that limit their use for certain applications.
For example, the aforementioned models operate in an offline manner, assuming the full clip (i.e. after the action has concluded) is available to make a decision.
Offline models are often inefficient in online settings, where action recognition must be performed based on the incomplete clip seen up until the current time. For example, the performance of Swin-B drops by $\sim$30\% on the Something-Something-v2 (SSv2) dataset when only the first 70\% frames are observed \cite{stergiou2022temporal}. 

Another common assumption is the requirement of complete spatial information over time. But, due to spatial redundancy, it is enough to observe only small but informative subregions of the full frames to make an accurate prediction. Several approaches~\cite{baradel2018glimpse, huang2022glance, mac2022efficient, wang2021adaptive, wang2022adafocus} primarily process narrow regions called ``glimpses''. 
However, these approaches still require the entire frame to determine informative glimpses.
While using a lightweight model for this ``global'' view reduces the overall computational cost, it still requires having a wide field of view initially, which does not come free. High-resolution, large FOV cameras are expensive, require more power, and consume more bandwidth to transmit data.
It is essential to minimize such costs in certain high-risk time-sensitive applications, such as mobile drones for disaster recovery, monitoring at-risk animals in the wild, or real-time translation of sign language.

We thus develop an inexpensive model that predicts informative glimpse locations \textit{without} observing whole frames, therefore obviating the need for high-resolution, large FOV cameras.
Starting from a glimpse at a given location, our model decides which location to attend to in subsequent frames solely based on previously observed glimpses. Consequently, our model predicts an action using only the local information and in an online fashion.
We choose transformers~\cite{vaswani2017attention} to learn glimpse-based attention mechanism and action prediction, as they can efficiently encode the relations between spatially and temporally distant glimpses. We thus call our model \textbf{Glimpse Transformers (\glitter)}. Following a \textit{factorized encoder} architecture \cite{arnab2021vivit}, we use a) a spatial encoder that solely models relations between the patches from a single glimpse to predict spatial features, and b) two temporal encoders that model interactions between various glimpse features across time to predict the class label and the next glimpse location, respectively. 

Since the ground truth for optimal glimpse locations is unavailable, we propose a novel spatiotemporal teacher-student consistency objective to incentivize \glitter to learn glimpse location in a weakly supervised manner. With only glimpses, \glitter (as the student model) is trained to reproduce the spatial features and class distribution of a teacher model ingesting the complete frames of the video.
As the teacher learns to produce predictive features and logits for the downstream task of online action recognition from the full frames, enforcing this consistency loss on the student model implicitly requires focusing attention on the most informative regions, leading to learning a glimpse mechanism.
We demonstrate \glitter's effectiveness on Something-Something-v2 \cite{goyal2017something} and Jester \cite{materzynska2019jester} datasets. Our main contributions are as follows.
\setlist{nolistsep}
\begin{itemize}[noitemsep]
\item We develop \glitter: an online action prediction model that observes only glimpses and predicts ongoing action based on partial spatiotemporal observations. While previous works locate glimpses by first observing full frames, \glitter predicts the next informative glimpse location solely based on the past glimpses.
\item We propose a novel spatiotemporal consistency objective to train \glitter without the ground truth for glimpse location. Under this objective, \glitter must select glimpses that summarize features and class distribution predicted from the entire frames. Our proposed consistency yields $\sim 10\%$ gain in accuracy on SSv2 compared to the baseline cross-entropy objective.
\item Our model that never observes complete frames and recognizes action solely based on local information gathered through glimpses achieves nearly 53\% and 94\% accuracy on SSv2 and Jester dataset, respectively, while reducing the total area observed per frame by nearly 67\% (with the glimpses of size 128$\times$128 extracted from frames of size 224$\times$224).
\end{itemize}

\noindent Our code can be found at: \url{https://github.com/facebookresearch/GliTr}.

\section{Related Works}
\minisection{Online Action Recognition.}
Many state-of-the-art methods perform offline action recognition once the entire video is available \cite{feichtenhofer2020x3d,feichtenhofer2019slowfast,carreira2017quo,kaufman2017temporal,tran2019video,tran2018closer,wang2016temporal,wang2018non}. However, these methods are not optimized for the case where the entire video is not yet available, and the model has to predict the action based on a preliminary, incomplete video.

Performing an online or early action recognition based on a partially observed video is a challenging task. A partially observed video may associate with multiple possible actions, leading to the inherent uncertainty in the prediction task. Several methods focus on predicting actions from partial videos. Zhao \etal~\cite{zhao2019spatiotemporal}, Wu \etal~\cite{wu2021anticipating}, and Pang \etal~\cite{pang2019dbdnet} anticipate future actions based on the motion and object relations in the past frames. Many analyze micro-motions in the available early frames \cite{stergiou2022temporal, lan2014hierarchical, kong2014discriminative, kong2015max}. Other approaches such as dynamic bag-of-words \cite{ryoo2011human}, global-local saliency~\cite{lai2017global}, memorizing hard-to-predict samples~\cite{kong2018action}, soft regression with multiple soft labels \cite{hu2018early}, and probabilistic modeling \cite{li2014prediction,cao2013recognize} are also used.
While the existing online action recognition methods focus on partial observation in the temporal dimension, we focus on partial information in the temporal as well as the spatial dimension.

\minisection{Spatial Selection for Action Recognition.}
Spatial selection is typically performed using hard attention \cite{mnih2014recurrent}. As opposed to soft attention models \cite{xu2015show} that observe all regions of the scene with varying attention level, hard attention models sequentially attend to the most informative glimpses. Hard attention is widely used for image classification \cite{ba2015multiple,elsayed2019saccader,mnih2014recurrent,xu2015show,  papadopoulos2021hard,rangrej2022consistency,rangrej2021probabilistic,wang2020glance}.

Recently, hard attention has also been applied to video action recognition. Wang \etal propose online action recognition model called Adafocus~\cite{wang2021adaptive,wang2022adafocus}. Chen \etal \cite{chen2020learning}, Huang \etal \cite{huang2022glance} and Wang \etal \cite{wang2017better} present offline models that first observe the entire video in order to predict attention-worthy glimpse locations. Mac \etal \cite{mac2022efficient} and Baradel \etal \cite{baradel2018glimpse} also present offline models but locate and observe multiple informative glimpses per-frame. Another line of approach leverages pose information and focuses only on the relevant body parts \cite{baradel2017human,das2019focus}. The previous approaches, irrespective of their online or offline nature, access full frames to locate informative glimpses. In contrast, our model never observes complete frames; it only observes a narrow glimpse from each frame.

\begin{figure*}
    \centering
    \includegraphics[page=1,trim={0 0 0 0},clip,width=\linewidth]{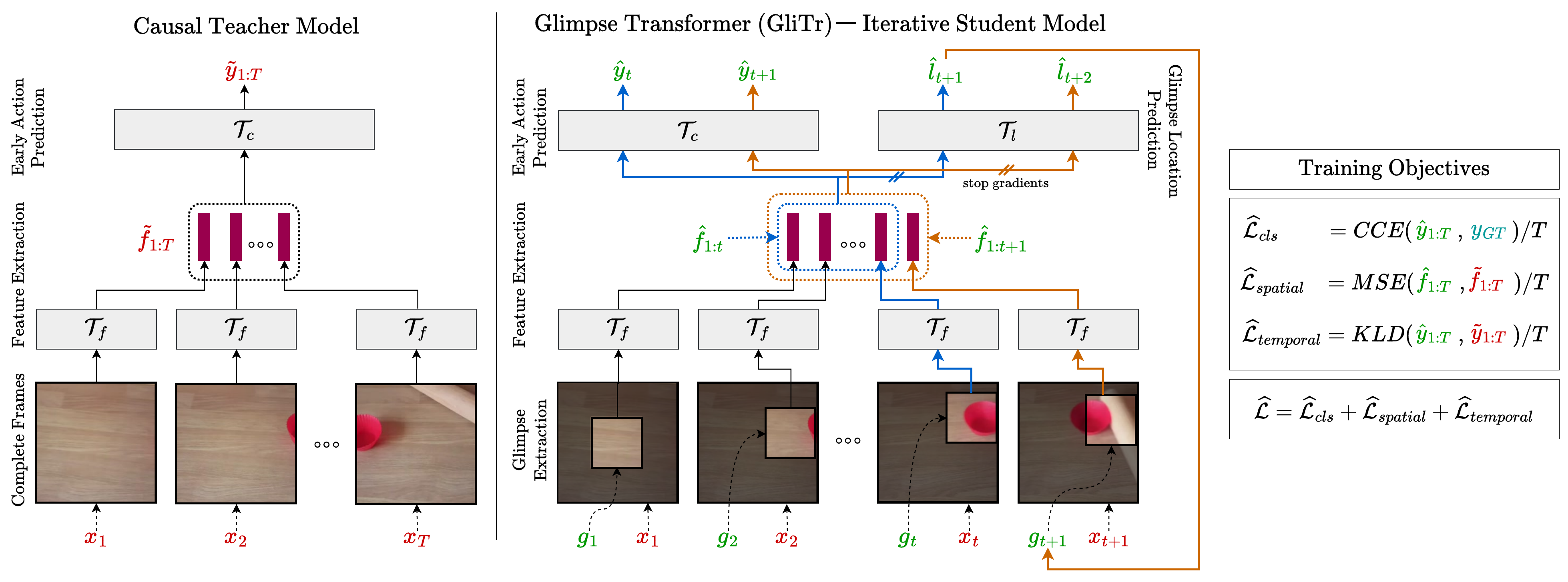}
    \caption{\textbf{An overview of our \glitter.} \glitter consists of a frame-level spatial transformer $\mathcal{T}_f$ and causal temporal transformers $\mathcal{T}_c$ and $\mathcal{T}_l$. One training iteration requires $T$ forward passes through our model. Above, we show two consecutive forward passes at time $t\leq T-1$ and $t+1\leq T$. \textbf{Forward pass t} ({\color{figblue}blue path}): Given a new glimpse $g_t$, $\mathcal{T}_f$ extracts glimpse-features $\hat{f}_t$. We append $\hat{f}_t$ to $\hat{f}_{1:t-1}$, \ie features extracted from $g_{1:t-1}$ during previous passes. Next, $\mathcal{T}_c$ predicts label $\hat{y}_t$ from $\hat{f}_{1:t}$. Simultaneously, $\mathcal{T}_l$ predicts next glimpse location $\hat{l}_{t+1}$ from $\hat{f}_{1:t}$. \textbf{Forward pass t+1} ({\color{figorange}orange path}): Given a predicted location $\hat{l}_{t+1}$, we extract a glimpse $g_{t+1}$ at $\hat{l}_{t+1}$ from a frame $x_{t+1}$. Then, we follow the same steps as the blue path. After $T$ forward passes, we compute the losses shown in the right. To find targets $\tilde{y}_{1:T}$ and $\tilde{f}_{1:T}$ for spatial and temporal consistency, we use a separate pre-trained and fixed teacher model (shown on the left and explained in Figure \ref{fig:teacher}) that observes complete frames $x_{1:T}$. To maintain stability, we stop gradients from $\mathcal{T}_l$ to $\mathcal{T}_f$.}
    \label{fig:student}
    \vspace{-4mm}
\end{figure*}
\minisection{Consistency Learning.}
Consistency is widely used for the problem of semi-supervised learning \cite{sajjadi2016regularization, sohn2020fixmatch, xie2020unsupervised, berthelot2019remixmatch, laine2016temporal}. The idea is to force the output of the model to be invariant to different augmentations of the same input \cite{sohn2020fixmatch, xie2020unsupervised, berthelot2019remixmatch, liang2021mixkd}, or variations in the internal representations \cite{bachman2014learning, sajjadi2016regularization}, or the model parameters at different training epochs \cite{laine2016temporal}. Another related approach is pseudo-labeling \cite{xie2020self, pham2021meta}, where a separate teacher model generates pseudo-labels for unlabeled samples under no perturbations, and the student model is trained to predict the pseudo-labels under some perturbations. This approach is similar to Knowledge Distillation \cite{hinton2015distilling}, where the student is trained to reconstruct the output or internal representation \cite{aguilar2020knowledge} of the teacher.

Many early action recognition models learn to predict the class distribution consistent with the complete video using only a subset of early frames \cite{cai2019action, fernando2021anticipating, kong2017deep, qin2017binary,wang2019progressive}. Others have also leveraged spatiotemporal consistency for complete frames \cite{tong2022videomae,feichtenhofer2022masked}. Inspired by previous work, we use a teacher model that predicts features from complete frames and predicts class distribution in an online fashion. Our student model observes partial spatiotemporal information and tries to predict features and class distribution consistent with the teacher model.

\section{Models}
We use a teacher model to i) initialize our \glitter~- a student model and ii) compute targets for the spatiotemporal consistency objective used for training \glitter. We discuss our teacher model in Sec \ref{sec:teacher} followed by \glitter in Sec \ref{sec:student}. We crown the quantities computed by our models using complete frames and glimpses with ($~\tilde{}~$) and ($~\hat{}~$), respectively.
\subsection{Teacher}
\label{sec:teacher}
Given spatially complete frames $x_{1:t}$ from a preliminary video at time $t\leq T$, our online teacher model predicts $\tilde{y}_t$, an early estimate of true action $y_{GT}$. We adapt \textit{factorized transformers encoder} architecture \cite{arnab2021vivit} for our teacher model, and aggregate spatial and temporal information sequentially. It includes the following components.

\minisection{Feature Extraction ($\mathcal{T}_f$).} We use a spatial transformer $\mathcal{T}_f$ to extract features $\tilde{f}_{t}$ from each individual frame $x_t$ for all $t$. We use the ViT architecture \cite{vaswani2017attention,touvron2021training} without the final classification head and collect features from the output corresponding to the input class token.

\minisection{Early Action Prediction ($\mathcal{T}_c$).} We use a temporal transformer $\mathcal{T}_c$ to aggregate features $\tilde{f}_{1:t}$ and predict label $\tilde{y}_t$. Since transformers are permutation invariant, we enforce order in the input sequence using temporal position embeddings. Moreover, we do not use a separate class token and pass the output corresponding to $\tilde{f}_{t}$ to the linear classifier to predict $\tilde{y}_t$. Further, to reduce training time, we use causal attention masking~\cite{girdhar2021anticipative, conneau2019cross}. Hence, during training, $\mathcal{T}_c$ observes $\tilde{f}_{1:T}$ and produces $\tilde{y}_{1:T}$ in a single forward pass while aggregating features in an online progressive manner, referencing only $\tilde{f}_{1:t}$ to produce output $\tilde{y}_t$ at index $t$.

\minisection{Glimpse Location Prediction ($\mathcal{T}_l$).} We include temporal transformer $\mathcal{T}_l$ to predict glimpse location $\tilde{l}_{t+1}$ from $\tilde{f}_{1:t}$. $\mathcal{T}_l$ has the same architecture as $\mathcal{T}_c$, except the final linear classifier is replaced by a linear regression head to predict coordinates $\tilde{l}_{t+1}$. Though not required for online action prediction from full frames, we train $\mathcal{T}_l$ to initialize the corresponding module in our student model. Once the student model is initialized, we discard $\mathcal{T}_l$ from the teacher model.

\subsection{Glimpse Transformer (\glitter) --- Student}
\label{sec:student}
Our Glimpse Transformer (\glitter) is derived and initialized from the teacher model discussed in Sec \ref{sec:teacher}. It is an iterative model that actively locates and attends to narrow glimpses in a scene and predicts an ongoing action early based on spatially and temporally incomplete observations. At time $t$, \glitter senses a new glimpse $g_t$ at location $\hat{l}_t$ from frame $x_t$. Using glimpses $g_{1:t}$, it predicts i) $\hat{y}_t$, an early approximation of label $y_{GT}$ and ii) $\hat{l}_{t+1}$, location of the next glimpse. We display schematics of \glitter in Figure \ref{fig:intro}. We illustrate \glitter's operation in Algorithm \ref{algo:inference} and Figure \ref{fig:student}. It consists of the following components.

\minisection{Glimpse Extraction.} Given a location $\hat{l}_t = (i,j)$, we crop a glimpse $g_t$ centered at location $\hat{l}_t$ in frame $x_t$. To maintain differentiability through the cropping operation, we use a spatial transformer network (STN) \cite{jaderberg2015spatial}\footnote{Not to be confused with (spatial) Vision Transformers (ViT) \cite{dosovitskiy2020image}.}.

\minisection{Feature Extraction ($\mathcal{T}_f$).} 
Similar to the teacher model, we use $\mathcal{T}_f$ to extract features $\hat{f}_t$ from glimpse $g_t$. We derive position embeddings for patches in $g_t$ using STN.

\minisection{Early Action Prediction ($\mathcal{T}_c$).} We input glimpse features $\hat{f}_{1:t}$ to $\mathcal{T}_c$ which in turn predicts class label $\hat{y}_t$. 

\minisection{Glimpse Location Prediction ($\mathcal{T}_l$).} Similarly, we pass the features $\hat{f}_{1:t}$ to $\mathcal{T}_l$ which predicts next glimpse location $\hat{l}_{t+1}$.

\setlength{\textfloatsep}{5pt}
\begin{algorithm}[t]
\small
  \caption{Inference using \glitter}
  \begin{algorithmic}[1]
      \State $\hat{l}_1$ is predefined.
      \For{$t \in \{1,\dots,T\}$}
        \State Sample $g_t$ at $\hat{l}_t$ from $x_t$. \Comment{\textbf{Glimpse Extraction}}
        \State $\hat{f}_t = \mathcal{T}_f(g_t,\hat{l}_t)$ \Comment{\textbf{Feature Extraction}}
        \State $\hat{y}_t = \mathcal{T}_c(\hat{f}_{1:t})$ \Comment{\textbf{Early Action Prediction}}
        \State $\hat{l}_{t+1} = \mathcal{T}_l(\hat{f}_{1:t})$ \Comment{\textbf{Glimpse Location Prediction}}
        \State Save $\hat{f}_t$.
      \EndFor
  \end{algorithmic}
  \label{algo:inference}
\end{algorithm}

\section{Training Objectives}
We discuss training objectives for \glitter in Sec \ref{sec:student_training}. Considering \glitter as the downstream model, we design training objectives suitable for our teacher model in Sec \ref{sec:teacher_training}. We crown training objectives of \glitter and the teacher model with ($~~\widehat{}~~$) and ($~~\widetilde{}~~$), respectively.

\subsection{Glimpse Transformer (\glitter) --- Student}
\label{sec:student_training}
\minisection{Classification Loss.} Since our goal is to predict action label $y_{GT}$ early using the spatially and temporally incomplete video, we minimize the cross-entropy loss given by
\begin{align}
    \widehat{\mathcal{L}}_{cls} = CCE(\hat{y}_{1:T}, y_{GT})/T.
\end{align}

\minisection{Spatial Consistency Loss.} We require \glitter to attend to the glimpses that produce features as predictive of the action as the ones predicted using complete frames by our teacher model. Hence, we minimize the mean squared error (MSE) between the glimpse features $\hat{f}_{t}$ predicted by \glitter and the frame features $\tilde{f}_t$ predicted by our teacher model, which is
\begin{align}
    \widehat{\mathcal{L}}_{spatial} = MSE(\hat{f}_{1:T}, \tilde{f}_{1:T})/T.
    \label{eq:student_spatio_consistency}
\end{align}

\minisection{Temporal Consistency Loss.}
While the teacher model has all instantaneous spatial information available in a complete frame, \glitter must rely on past glimpses to reason about the unobserved yet informative regions in the current frame. To incentivize \glitter to aggregate spatial information from the past to mitigate partial observability, we minimize the KL-divergence between the class logits predicted by \glitter using glimpses ($\hat{y}_t$) and the teacher using complete frames ($\tilde{y}_t$), yielding
\begin{align}
    \widehat{\mathcal{L}}_{temporal} = KLD(\hat{y}_{1:T}, \tilde{y}_{1:T})/T.
    \label{eq:student_temporal_consistency}
\end{align}
Our final training objective for GliTr is the following:
\begin{align}
    \widehat{\mathcal{L}} = \widehat{\mathcal{L}}_{cls} + \widehat{\mathcal{L}}_{spatial} + \widehat{\mathcal{L}}_{temporal}
\end{align}

\subsection{Teacher}
\label{sec:teacher_training}
\begin{figure*}
    \centering
    \includegraphics[page=1,trim={0 0 0 0},clip,width=\linewidth]{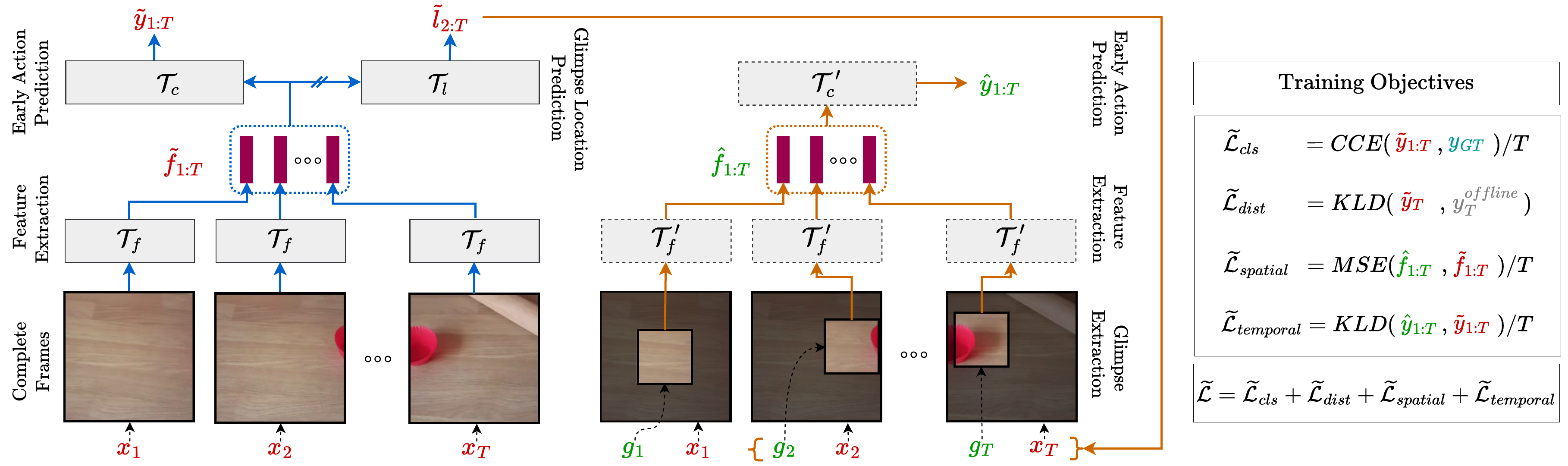}
    \caption{\textbf{An overview of our teacher model.} Our teacher model consists of a spatial transformer $\mathcal{T}_f$ and \textit{causal} temporal transformers $\mathcal{T}_c$ and $\mathcal{T}_l$. Each training iteration of the teacher model consists of two steps. \textbf{Step 1} ({\color{figblue}blue path}): Given complete video frames $x_{1:T}$, $\mathcal{T}_f$ extracts frame features $\tilde{f}_{1:T}$. Next, $\mathcal{T}_c$ and $\mathcal{T}_l$ predict class labels $\tilde{y}_{1:T}$ and glimpse locations $\tilde{l}_{2:T+1}$ from $\tilde{f}_{1:T}$, respectively. We discard $\tilde{l}_{T+1}$. \textbf{Step 2} ({\color{figorange}orange path}): Given $\tilde{l}_1$ (learnable parameter) and $l_{2:T}$ (predicted in step 1), we extract glimpses $g_{1:T}$ from $x_{1:T}$. Then, we create non-learnable copies of $\mathcal{T}_f$ and $\mathcal{T}_c$ denoted as $\mathcal{T}'_f$ and $\mathcal{T}'_c$. $\mathcal{T}'_f$ extracts glimpse-features $\hat{f}_{1:T}$ from $g_{1:T}$ and $\mathcal{T}'_c$ predicts labels $\hat{y}_{1:T}$ from $\hat{f}_{1:T}$. We compute losses shown on the right and update model parameters. To achieve stability during training, we stop gradients from $\mathcal{T}_l$ to $\mathcal{T}_f$.}
    \label{fig:teacher}
    \vspace{-4mm}
\end{figure*}
\minisection{Classification loss.} For all $t$, we minimize cross-entropy loss between the prediction $\tilde{y}_t$ and the ground-truth label $y_{GT}$ of the action,
\begin{align}
    \widetilde{\mathcal{L}}_{cls} = CCE(\tilde{y}_{1:T}, y_{GT})/T.
\end{align}

\minisection{Distillation loss.} When available, we also use a more powerful offline action recognition model such as VideoMAE~\cite{tong2022videomae} to predict action $y^{offline}_T$ from a complete video, \ie $x_{1:T}$. Then, we minimize the KL-divergence between the final prediction $\tilde{y}_T$ and the above $y^{offline}_T$ given by
\begin{align}
    \widetilde{\mathcal{L}}_{dist} = KLD(\tilde{y}_T, y^{offline}_T).
\end{align}

\minisection{Spatiotemporal Consistency losses.} Note that the above two losses train only $\mathcal{T}_f$ and $\mathcal{T}_c$. We use the following strategy to train $\mathcal{T}_l$. First, we use the locations $\tilde{l}_1$ (learnable parameter) and $\tilde{l}_{2:T}$ predicted by $\mathcal{T}_l$, to extract glimpses $g_{1:T}$ from frames $x_{1:T}$. Next, we create copies of $\mathcal{T}_f$ and $\mathcal{T}_c$ denoted as $\mathcal{T}'_f$ and $\mathcal{T}'_c$. We input $g_{1:T}$ and the corresponding position embeddings to $\mathcal{T}'_f$ and predict glimpse features $\hat{f}_{1:T}$. Given $\hat{f}_{1:T}$, $\mathcal{T}'_c$ predicts actions $\hat{y}_{1:T}$ in an online fashion. Then we minimize,
\begin{align}
    \widetilde{\mathcal{L}}_{spatial} = MSE(\hat{f}_{1:T}, \tilde{f}_{1:T})/T, \\
    \widetilde{\mathcal{L}}_{temporal} = KLD(\hat{y}_{1:T}, \tilde{y}_{1:T})/T.
    \label{eq:teacher_consistency}
\end{align}
We use the above two losses to update parameters of $\mathcal{T}_l$ only. We design these consistency objectives based on the spatiotemporal consistency objectives of \glitter (equations \ref{eq:student_spatio_consistency} and \ref{eq:student_temporal_consistency}). As discussed in Sec \ref{sec:student_training}, they encourage $\mathcal{T}_l$ to locate glimpses covering the most useful task-relevant regions in the frames, but based on complete frames observed in the past. We demonstrate the training procedure in Figure \ref{fig:teacher}.

The final objective for our teacher model is as follows. 
\begin{align}
    \widetilde{\mathcal{L}} = \widetilde{\mathcal{L}}_{cls} + \widetilde{\mathcal{L}}_{dist} +\widetilde{\mathcal{L}}_{spatial} + \widetilde{\mathcal{L}}_{temporal}
\end{align}

\section{Experiments}
\label{sec:implementation}
\begin{figure*}
    \centering
    \begin{minipage}{0.65\linewidth}
    \vspace{-1.5mm}
    \includegraphics[page=1,trim={5cm 0 5cm 0},clip,width=\textwidth]{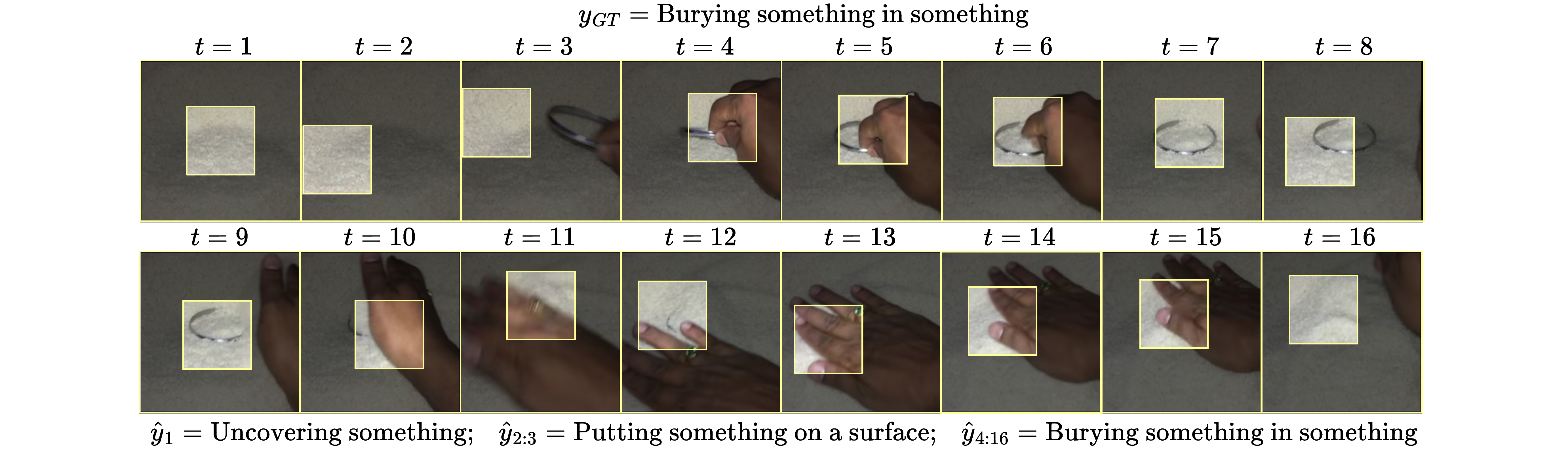}
    \end{minipage}
    \hfill
    \begin{minipage}{0.33\linewidth}
      \includegraphics[width=\textwidth]{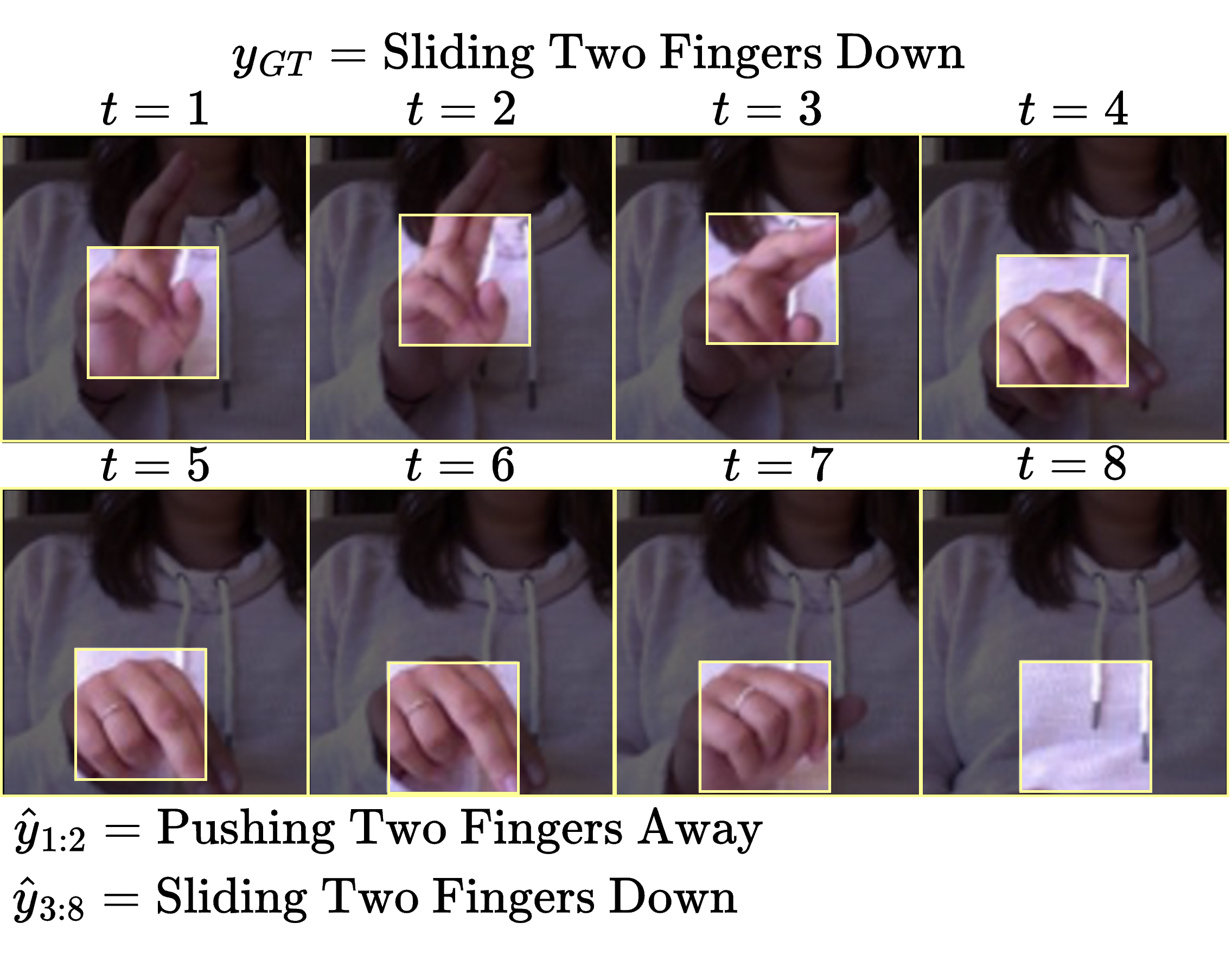}
    \end{minipage}
    \vspace{-4mm}
    \caption{\textbf{Glimpses selected by \glitter} on (left) SSv2 and (right) Jester. The complete frames are shown for reference only. \glitter does not observe full frames. It only observes glimpses. We show additional examples in Appendix \ref{sec:appendix}.}
    \label{fig:visualization}
    \vspace{-4mm}
\end{figure*}

\minisection{Datasets.} We experiment with two publicly available large-scale real-world datasets, namely, Something-Something-v2 (SSv2) \cite{goyal2017something} and Jester \cite{materzynska2019jester}. We adopt the official training-validation splits. SSv2 dataset contains videos recording 174 human actions using everyday objects. There are $\sim$170K videos for training and $\sim$25K for validation. Jester dataset is a collection of videos capturing 27 basic hand gestures, consisting of $\sim$120K videos for training and $\sim$15K videos for validation.

\minisection{Implementation.} We sample a sequence of 16 and 8 frames per video from SSv2 and Jester, respectively. We resize each frame to size $224\times224$ and use glimpses of size $96\times96$, unless stated otherwise. We use ViT-Small \cite{touvron2021training} architecture for $\mathcal{T}_f$. For $\mathcal{T}_c$ and $\mathcal{T}_l$, we use a custom transformers architecture with 768 embedding dimensions, 6 heads, and a depth of 4. 

\minisection{Optimization.} First, we discuss the common setting followed by a model-specific setting. For all models and datasets, we use the same data augmentation scheme as the one used for VideoMAE \cite{tong2022videomae}. Similar to Wang \etal \cite{wang2022adafocus}, we stop gradients from $\mathcal{T}_l$ to $\mathcal{T}_f$ to maintain stability during training. We use AdamW optimizer~\cite{loshchilov2018decoupled} with weight decay of 5e-2 and cosine learning rate schedule with no warmup unless stated otherwise. We run experiments for SSv2 and Jester on 4 A100 GPUs with 40 GB of memory and 4 V100-SXM2 GPUs with 32 GB of memory, respectively.

To train a teacher model on SSv2 dataset, we initialize $\mathcal{T}_f$ using an open-source ViT-S model~\cite{zhou2021image} pretrained on the ImageNet dataset~\cite{deng2009imagenet}, and initialize $\mathcal{T}_c$ and $\mathcal{T}_l$ randomly. We form a mini-batch using $b=60$ videos and use an initial learning rate of $\frac{\alpha b}{128}$, with base learning rate $\alpha$ being 1e-5, 1e-4 and 1e-4 for $\mathcal{T}_f$, $\mathcal{T}_c$ and $\mathcal{T}_l$, respectively. We train the teacher model for 40 epochs with a warmup of 15 epochs for $\mathcal{T}_l$. 
For the Jester dataset, we initialize the teacher model with the teacher model trained on the SSv2 dataset. We do not use distillation loss $\widetilde{\mathcal{L}}_{dist}$ for Jester dataset. We use a batch size $b$ of 100 and $\alpha$ of 1e-5 for all modules. The model is trained for 50 epochs. 

Each student model (\glitter) is initialized from a teacher model trained on the corresponding dataset. We use base learning rate $\alpha=$ 1e-5 for all modules and train them for 100 and 150 epochs with a batch-size b of 360 and 800 videos from SSv2 and Jester, respectively.

\subsection{Empirical Comparisons}
\minisection{Glimpse Mechanisms Under Partial Observability}

We compare the glimpse attention strategy learned by \glitter with four baselines and an approximate upper bound:
\begin{tight_itemize}
    \item{\em Uniform random}: Glimpse locations are independently drawn from a uniform distribution for each $t$. 
    \item{\em Gaussian random}: Similar to \textit{uniform random} but instead, the glimpse locations are sampled from Gaussian distribution with zero mean and unit variance and passed through a $\mathrm{tanh()}$ function to constrain locations to remain within the bounds of the frame.
    \item{\em Center}: The model observes glimpses from a constant location at the center of each frame.
    \item{\em Bottom Left}: The model attends to the glimpses in the bottom left corner of the frames.
    \item{\em Teacher (an upper bound)}: Glimpse locations are chosen as predicted by the teacher model which looks at the full frames. In the absence of ground truth glimpse locations, this provides an approximate upper bound.
\end{tight_itemize}
To isolate the glimpse strategy's effect on performance, we evaluate the glimpses selected by various strategies using the same model \ie \glitter. While assessing the baselines and the upper bound, we ignore predictions from $\mathcal{T}_l$ and instead use locations given by the specific strategies described above.
\begin{figure}
    \centering
    \begin{minipage}{0.58\linewidth}
      \includegraphics[width=\textwidth]{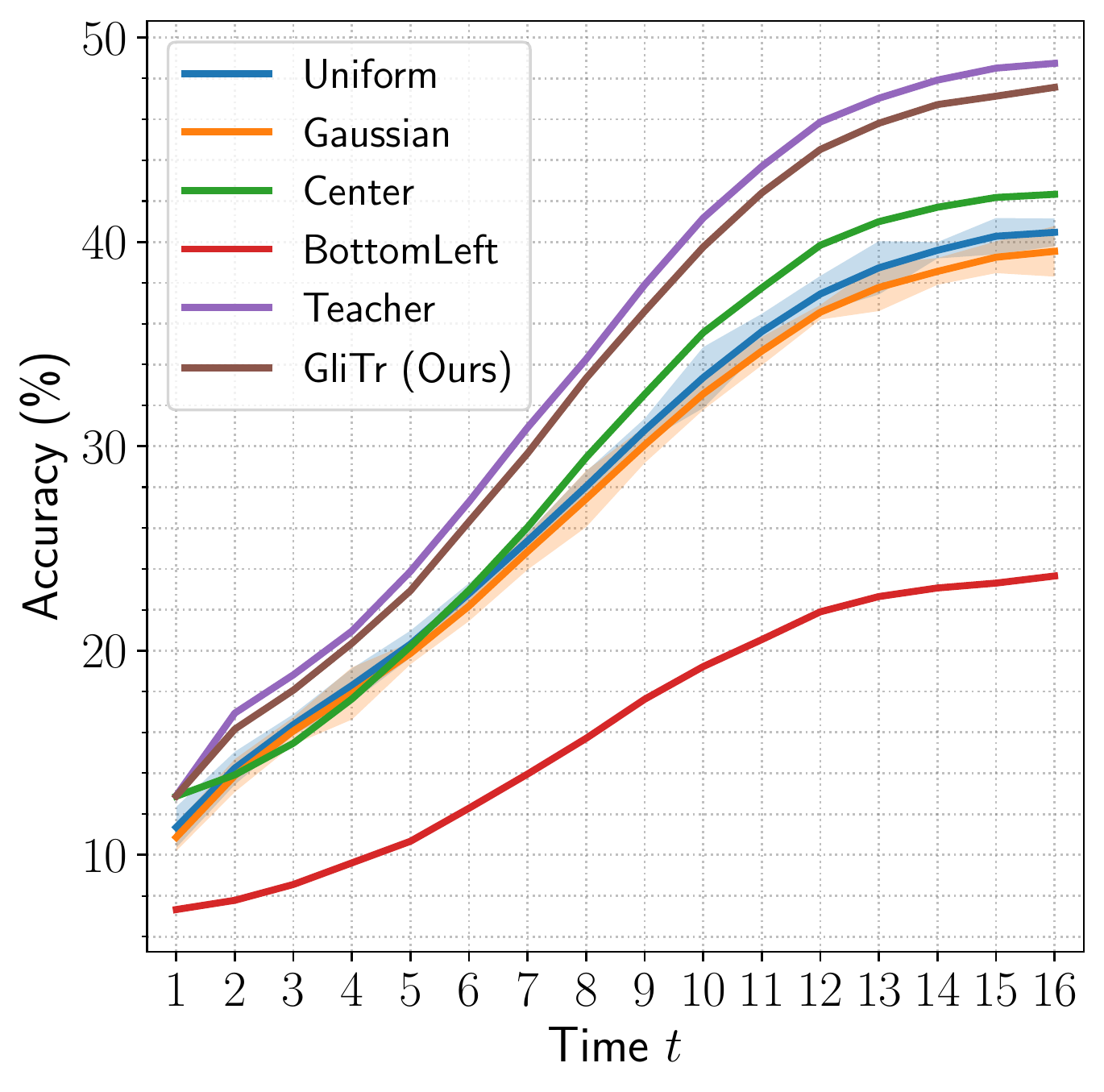}
      \par\vspace{-1.5mm}
      \centering{(a)}
    \end{minipage}
    \hfill
    \begin{minipage}{0.41\linewidth}
      \includegraphics[width=\textwidth]{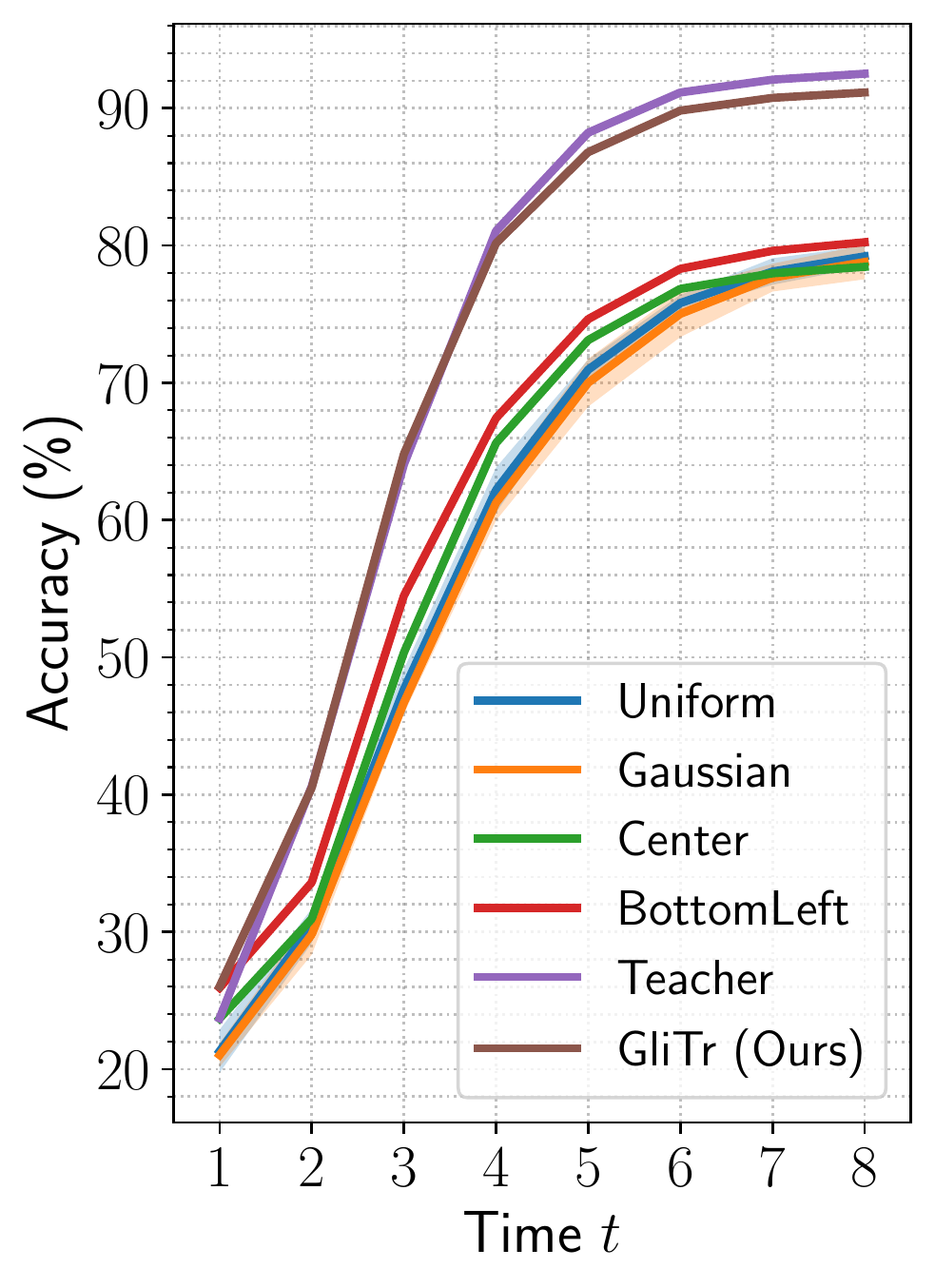}
      \par\vspace{-1.5mm}
      \centering{(b)}
    \end{minipage}
    \par\vspace{-2mm}
    \caption{\textbf{Comparison of online action prediction accuracy using different glimpse mechanisms.} (a) SSv2 and (b) Jester. The \textit{Uniform} and the \textit{Gaussian} strategies sample locations from the respective distributions. We display mean$\pm5\times$std computed using five independent runs. The \textit{Center} and the \textit{Bottom Left} strategies always observe glimpses at the constant locations. The \textit{Teacher} (an approximate upper bound) and our \glitter locate informative glimpses based on past frames and glimpses, respectively.}
    \label{fig:baseline}
\end{figure}
We show results in Figure \ref{fig:baseline}, plotting online action prediction accuracy after each $t$. As expected, the prediction accuracy for all strategies increases as the model observes more glimpses. The \textit{Center} and the \textit{Bottom Left} strategies outperform other baselines on SSv2 and Jester datasets, respectively. We suspect this is because the object of interest frequently appears in the center in SSv2; while in most examples from Jester, hand movements begin and end in the region near the bottom left corner of the frames. On the other hand, \glitter outperforms all baselines and achieves performance closest to the upper bound (\ie the \textit{Teacher} strategy). We plot a histogram of glimpse regions selected by \glitter in Figure \ref{fig:glimpse_hist}. We observe that not only does \glitter successfully capture different biases (center vs. bottom left) in the two datasets, but it also ignores the bias if necessary. Notice the spread in the histograms for $t>1$, suggesting \glitter observes various regions in different videos. Consequently, \glitter achieves better accuracy faster than the baselines, and at time $T$, outperforms the best performing baselines with the respective margins of nearly $5\%$ and $11\%$ on SSv2 and Jester. We visualize glimpses selected by \glitter on example videos from SSv2 and Jester in Figure \ref{fig:visualization}.

\begin{figure}
    \centering
    \begin{minipage}{0.65\linewidth}
      \includegraphics[width=\textwidth]{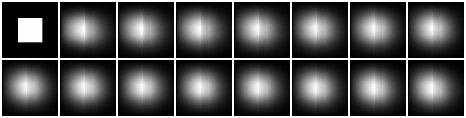}
      \centering{(a)}
    \end{minipage}
    \hfill
    \begin{minipage}{0.32\linewidth}
      \includegraphics[width=\textwidth]{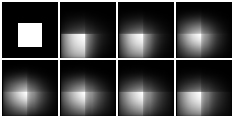}
      \centering{(b)}
    \end{minipage}
    \vspace{-2mm}
    \caption{\textbf{Histograms of the glimpse regions selected by \glitter} with increasing time (raster scan order) on (a) SSv2 and (b) Jester. Recall that \glitter observes the first glimpse at a predetermined location followed by active selection.}
    \vspace{3mm}
    \label{fig:glimpse_hist}
\end{figure}

\begin{table*}[!t]
  \centering
  \resizebox{\textwidth}{!}{%
  \begin{tabular}{lll|ccrr|ccrr} 
  \toprule[1.2pt]
  \textbf{Method} &	\textbf{Online/}  &	\textbf{Observes}     &  \multicolumn{4}{c}{\textbf{SSv2}\cite{goyal2017something}}	&  \multicolumn{4}{c}{\textbf{Jester}\cite{materzynska2019jester}}\\
                  &	\textbf{Offline?} &	\textbf{full frames?} & \textbf{Glimpse-size} &	\textbf{\#frames} & \textbf{\#pixels} & \textbf{Accuracy (\%)} & \textbf{Glimpse-size} &	\textbf{\#frames} & \textbf{\#pixels} & \textbf{Accuracy (\%)}\\
  \midrule[0.8pt]
AdaFocus \cite{wang2021adaptive}$^\diamond$   & Offline & Yes & 144$\times$144 & 8+12 & 1M & ~~~59.70 & - & - & - & -\\
 	                               &         &     & 160$\times$160 & 8+12 & 1M & 60.20 & - & - & - & -\\
 	                               &         &     & 176$\times$176 & 8+12 & 1M & 60.70 & - & - & - & -\\
AdaFocusV2 \cite{wang2022adafocus}$^\diamond$ & Offline & Yes & 128$\times$128 & 8+12	& 1M & 59.60 & 128$\times$128 & 8+12	& 1M & 96.60	\\
                                   &	     &     & 144$\times$144 & 8+12	& 1M & 60.50 & 176$\times$176 & 8+12 & 1M & 96.90	\\
	                               &	     &     & 160$\times$160 & 8+12 & 1M & 60.80 & - & - & - & - \\
	                               &	     &	   & 176$\times$176 & 8+12 & 1M & 61.30 & - & - & - & - \\
GFNet \cite{huang2022glance}$^\S$       & Offline & Yes & ~~~~~~~~~~96$\times$96 ($\times$2)$^\star$	& 8	& 401K & 59.50 & ~~~~~~~~~~80$\times$80 ($\times$2)$^\star$	& 8	& 401K & 95.50\\
                                   &         &     & ~~~~~~~~~~96$\times$96 ($\times$2)$^\star$	& 12& 602K & 61.00 & ~~~~~~~~~~96$\times$96 ($\times$2)$^\star$	& 12& 602K & 95.80\\
                                   &         &     & ~~~~~~~~~~96$\times$96 ($\times$2)$^\star$	& 16& 803K & 62.00 & ~~~~~~~~~~128$\times$128 ($\times$2)$^\star$	& 16& 803K & 96.10\\
\bottomrule
\glitter (Ours)	                    &\textbf{Online}&\textbf{No}&64$\times$64 & 16 &	\textbf{66K} & ~~~38.24	& 64$\times$64 & 8 & \textbf{33K} & 84.03	\\
 	                                &	         	&           &96$\times$96	& 16 &	\textbf{147K}& ~~~47.56	& 96$\times$96	& 8 & \textbf{74K}& 91.15	\\
 	                                &	        	&           &128$\times$128& 16 &	\textbf{262K}& ~~~53.02	&128$\times$128& 8 & \textbf{131K}& 93.91	\\
\bottomrule[1.2pt]
\end{tabular}
}
\vspace{-2mm}
\caption{\textbf{Comparison with glimpse-based action recognition models.} We count the number of pixels sensed by different approaches to perform recognition. Previous approaches are offline and use complete frames to locate informative glimpses and to recognize actions. \glitter is an online model and only observes glimpses, not complete frames. \glitter achieves competitive performance with a significant saving in the total area observed. $^\diamond$AdaFocus \cite{wang2021adaptive} and AdaFocusV2 \cite{wang2022adafocus} first observe 8 frames to locate useful glimpses and then sample additional 12 frames to extract glimpses, which requires sensing total 20 frames in advance due to their offline nature. $^\S$Results are based on Figure 13 from \cite{huang2022glance}. $^\star$GFNet observes two glimpses per frame. For comparison with online methods, refer to Figure \ref{fig:sota}.}
\vspace{-4mm}
\label{tab:sota}
\end{table*}

\minisection{Models with Complete Spatial Observability}

\minisection{Glimpse-based offline models.} We compare our \glitter with previous glimpse-based offline action recognition models in Table \ref{tab:sota}. We note that a direct comparison between these approaches is unfair since previous models also observe complete frames. Further, unlike offline approaches that initially observe a complete video and select an informative glimpse at $t$ based on the current, past, and future frames, our \glitter~- an online model - relies only on the past information to locate glimpses in the current frame. Moreover, previous methods use global information gathered from complete frames to locate glimpses and predict actions; however, \glitter only uses local information. Nevertheless, we include this analysis to highlight the savings achieved by \glitter in terms of the amount of area observed for recognition while still achieving competitive performance with partial observations.

We calculate and compare the number of pixels sensed by various methods to perform action recognition. AdaFocus \cite{wang2021adaptive} and AdaFocusV2 \cite{wang2022adafocus} uniformly sample 8 frames from a complete video to predict glimpse locations, followed by uniform sampling of another 12 frames to extract glimpses. In total, they require sensing 20 complete frames (20$\times$(224$\times$224)$\approx$1M pixels) in advance due to their offline nature. GFNet \cite{huang2022glance}, on the other hand, locates and extracts glimpses from the same set of complete frames. When compared to AdaFocusV2 with glimpses of size $128\times 128$, our \glitter reduces the amount of sensing by nearly $74\%$ and $87\%$ while compromising only around 6\% and 3\% accuracy on SSv2 and Jester, respectively. Further, while GFNet outperforms \glitter  by nearly 14.4\% and 4.7\% with glimpses of size 96$\times$96 on SSv2 and Jester, \glitter (with 16 and 8 glimpses, respectively) reduces the amount of sensing by nearly 82\% and 88\% compared to GFNet (with 16 and 12 frames, respectively) on these datasets. We emphasize that GFNet observes full frames and \textit{two} glimpses per frame in an offline manner, while \glitter observes only one glimpse per frame in an online fashion.

\begin{figure}
    \centering
    \begin{minipage}{0.57\linewidth}
      \includegraphics[width=\textwidth]{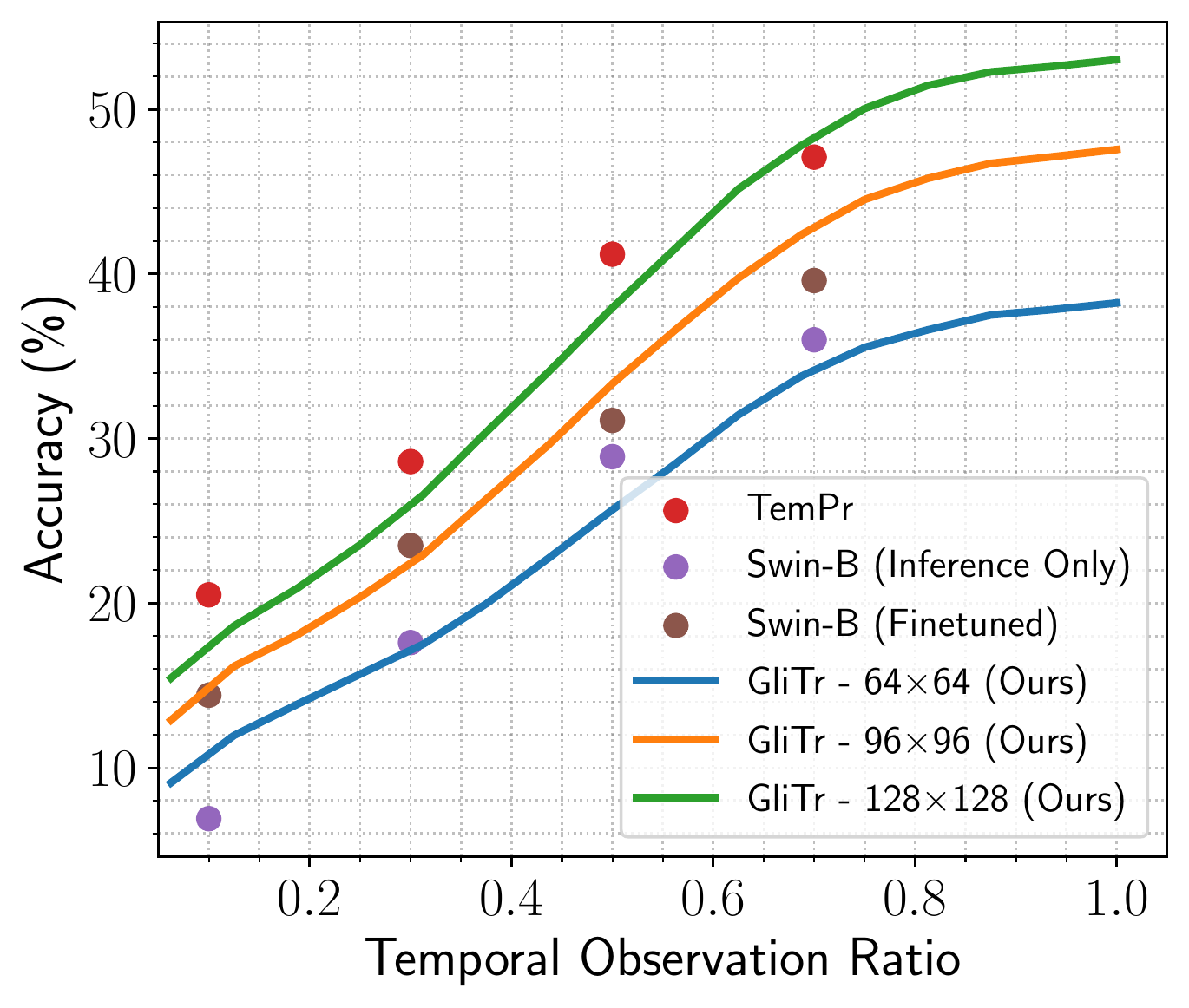}
      \vspace{-1mm}
      \centering{(a)}
    \end{minipage}
    \hfill
    \begin{minipage}{0.41\linewidth}
      \includegraphics[width=\textwidth]{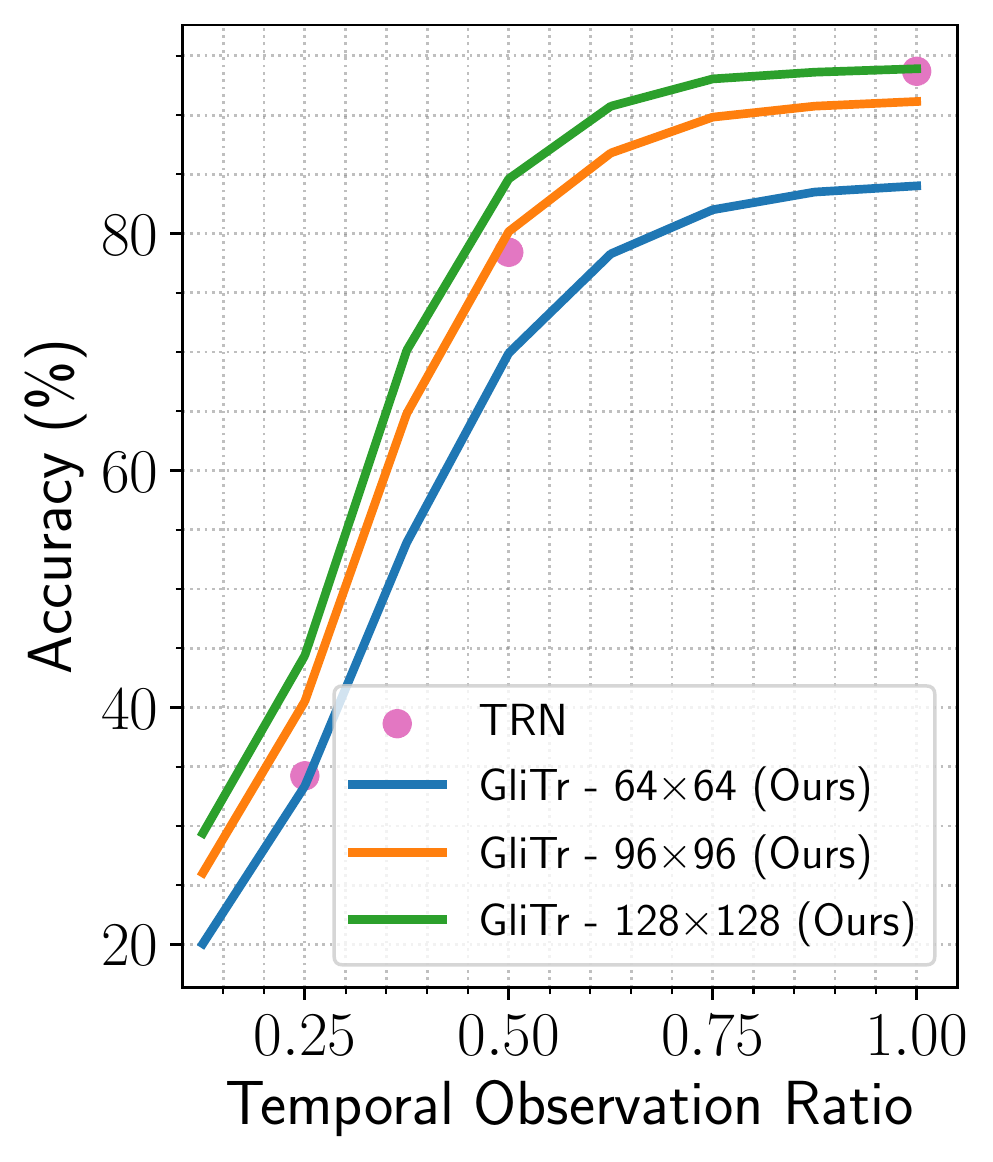}
      \vspace{-1mm}
      \centering{(b)}
    \end{minipage}
    \vspace{-1mm}
    \caption{\textbf{Comparison with early action prediction models.} (a) SSv2 and (b) Jester. While Swin-B~\cite{liu2022video}, TemPr \cite{stergiou2022temporal} and TRN~\cite{zhou2018temporal} predict action early based on complete frames, \glitter predicts action based on early glimpses.}
    \label{fig:sota}
    \vspace{2mm}
\end{figure}

\minisection{Early action prediction models}. We additionally compare \glitter with early action prediction models in Figure \ref{fig:sota}. We emphasize that these approaches observe entire frames (\ie global information) from a preliminary video; whereas, \glitter observes frames only partially through glimpses (\ie local information). For SSv2 dataset, we consider Swin-B~\cite{liu2022video} and TemPr \cite{stergiou2022temporal}. We cite Swin-B results from \cite{stergiou2022temporal}, who evaluate Swin-B for early action prediction before (\ie direct inference with pretrained model) and after finetuning on preliminary videos. Notice that, with glimpses of size $96 \times96$ and higher, \glitter outperforms Swin-B finetuned for early action prediction. Further, \glitter also outperforms TemPr with the glimpses of size $128 \times128$ when both have observed early 70\% video. For the Jester dataset,  \glitter outperforms TRN~\cite{zhou2018temporal} for early action prediction with glimpses of size $96 \times96$ and higher. The results demonstrate the efficiency of \glitter for early action prediction using only local information.

\subsection{Ablation on Spatiotemporal Consistency}
\begin{figure}
    \centering
    \includegraphics[page=1,trim={0.25cm 0 0 0},clip,width=\linewidth]{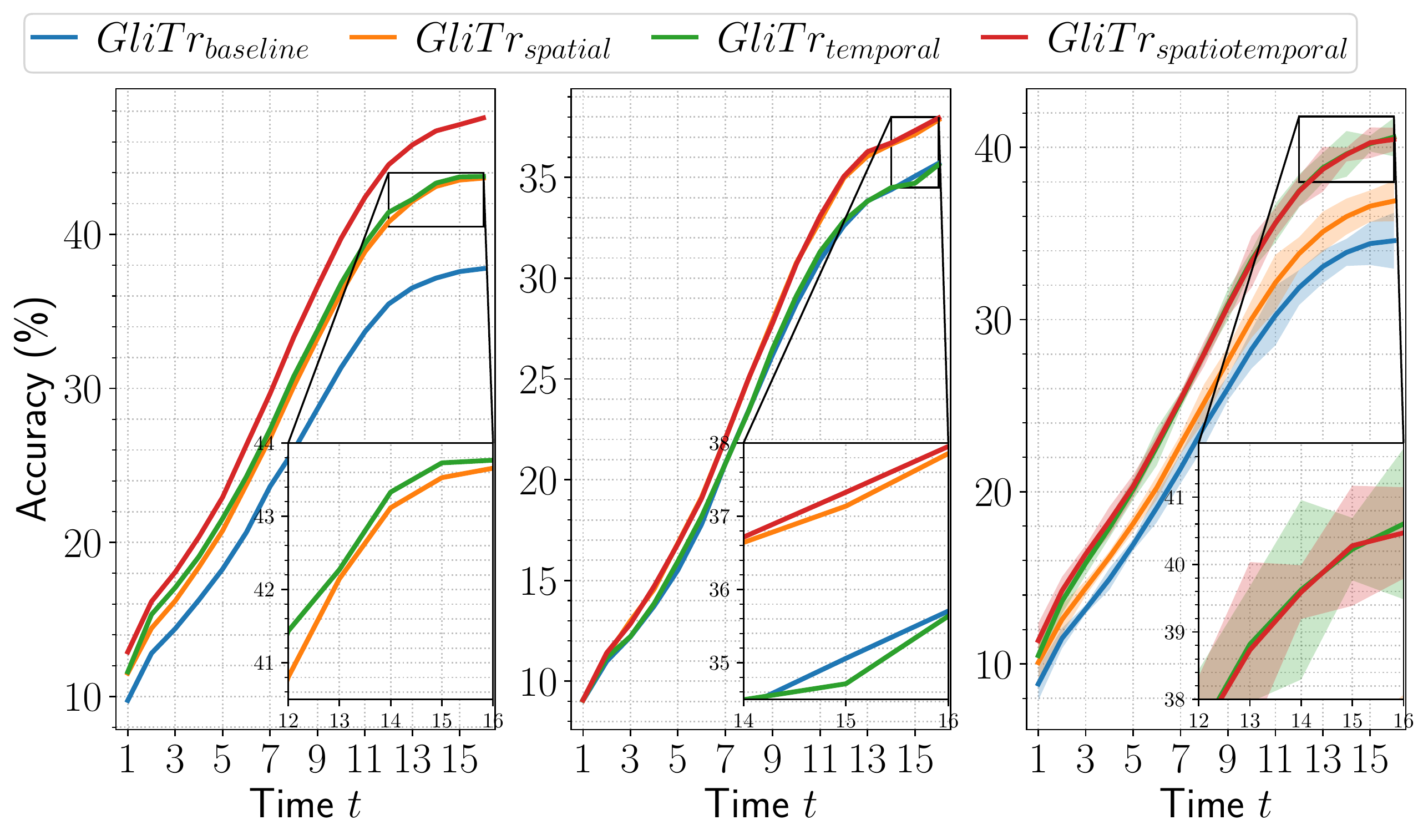}
    \begin{flushleft}
    \vspace{-3mm}
    ~~~~~~~~~~~~~~~~~(a)~~~~~~~~~~~~~~~~~~~~~~~~~~~(b)~~~~~~~~~~~~~~~~~~~~~~~~~~(c)
    \vspace{-4mm}
    \end{flushleft}
    \caption{\textbf{Ablation study on the spatiotemporal consistency objective} on SSv2 dataset. (a) accuracy of \glitter when trained using different combinations of the training objectives (b) accuracy of the teacher with the glimpses selected by the above variants. (c) accuracy of the above variants of \glitter when tested with the \textit{Uniform random} strategy. We display mean$\pm5\times$std from five independent runs.}
    \label{fig:loss_ablation}
\end{figure}

To demonstrate the value of the proposed spatiotemporal training objectives, we perform an ablation study for each on the SSv2 dataset. We train four variants of \glitter using the following combinations of the training objectives: i) $\text{\glitter}_{\text{baseline}}$ using $\widehat{\mathcal{L}}_{cls}$, ii) $\text{\glitter}_{\text{spatial}}$ using $\widehat{\mathcal{L}}_{cls}+\widehat{\mathcal{L}}_{spatial}$, iii) $\text{\glitter}_{\text{temporal}}$ using $\widehat{\mathcal{L}}_{cls}+\widehat{\mathcal{L}}_{temporal}$, and iv) our default variant $\text{\glitter}_{\text{spatiotemporal}}$ using $\widehat{\mathcal{L}}_{cls}+\widehat{\mathcal{L}}_{spatial}+\widehat{\mathcal{L}}_{temporal}$. Note that the above variants have the same architecture and operation; only their training objectives are different. Figure \ref{fig:loss_ablation}(a) shows results. We observe that including only spatial or only temporal consistency in the training objectives boosts \glitter's accuracy by nearly 6\% at $t$=16. Moreover, including both spatial and temporal consistency provides the highest improvement of around $10\%$.

To understand the sources of improvements provided by the two consistency losses, we perform two more experiments.
First, we evaluate glimpse selection strategies learnt by the above versions of \glitter using an impartial teacher model in Figure \ref{fig:loss_ablation}(b). We observe better performance for \glitter when spatial consistency is included in the training objectives, indicating that spatial consistency helps \glitter learn \textit{better glimpse selection strategy} and in turn improves its performance.
Second, we evaluate above four versions of \glitter using an impartial \textit{Uniform random} strategy in Figure \ref{fig:loss_ablation}(c). We observe that \glitter provides the highest performance for the \textit{Uniform random} strategy when we include temporal consistency in the training objective, suggesting that temporal consistency improves \glitter's performance by learning \textit{a better classifier under partial observability.} We experiment with different training procedures for the teacher model in Appendix \ref{sec:appendix}.

\subsection{Early Exit}
\begin{figure}
    \centering
    \begin{minipage}{0.58\linewidth}
      \includegraphics[width=\textwidth]{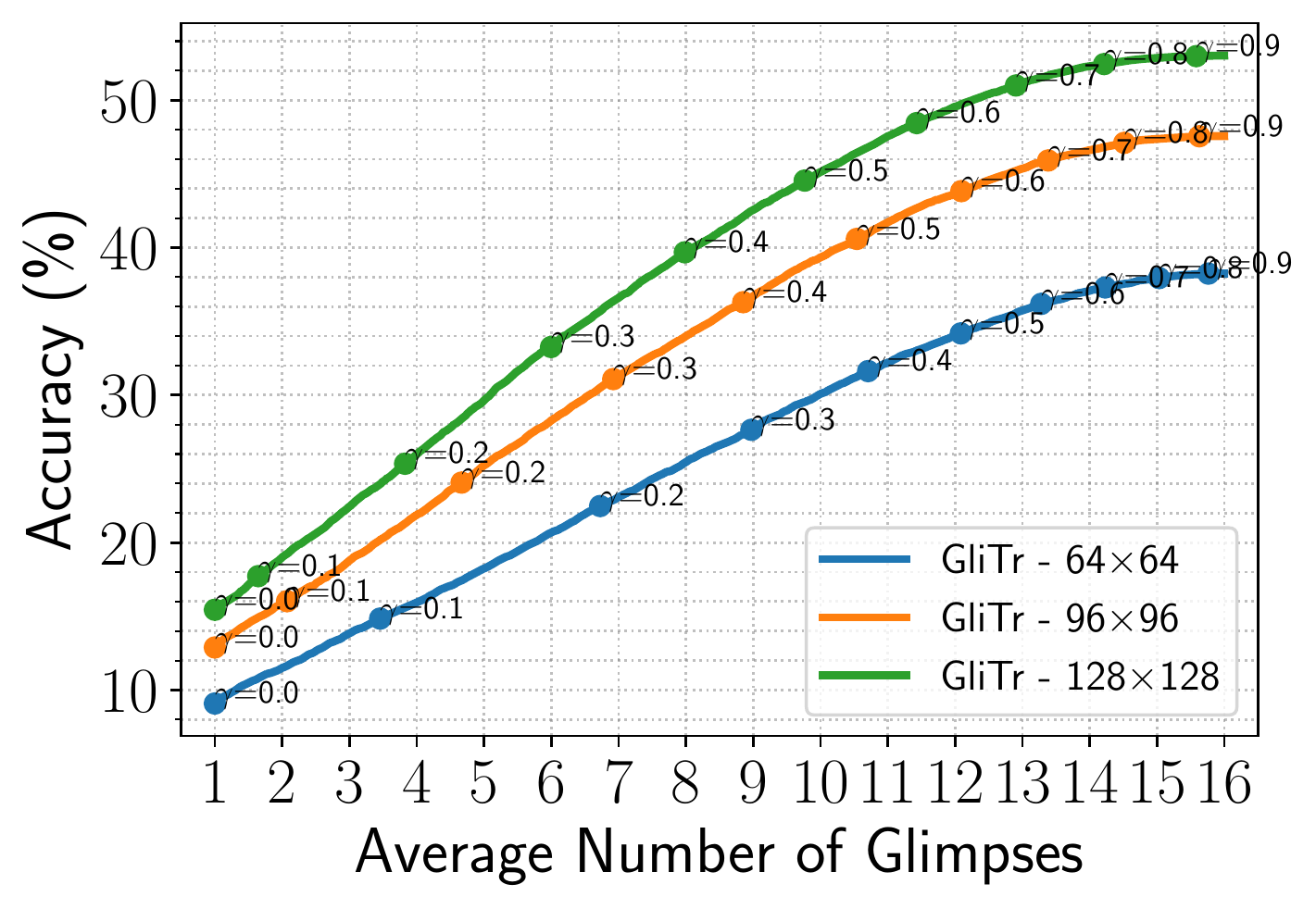}
      \centering{(a)}
    \end{minipage}
    \hfill
    \begin{minipage}{0.41\linewidth}
      \includegraphics[width=\textwidth]{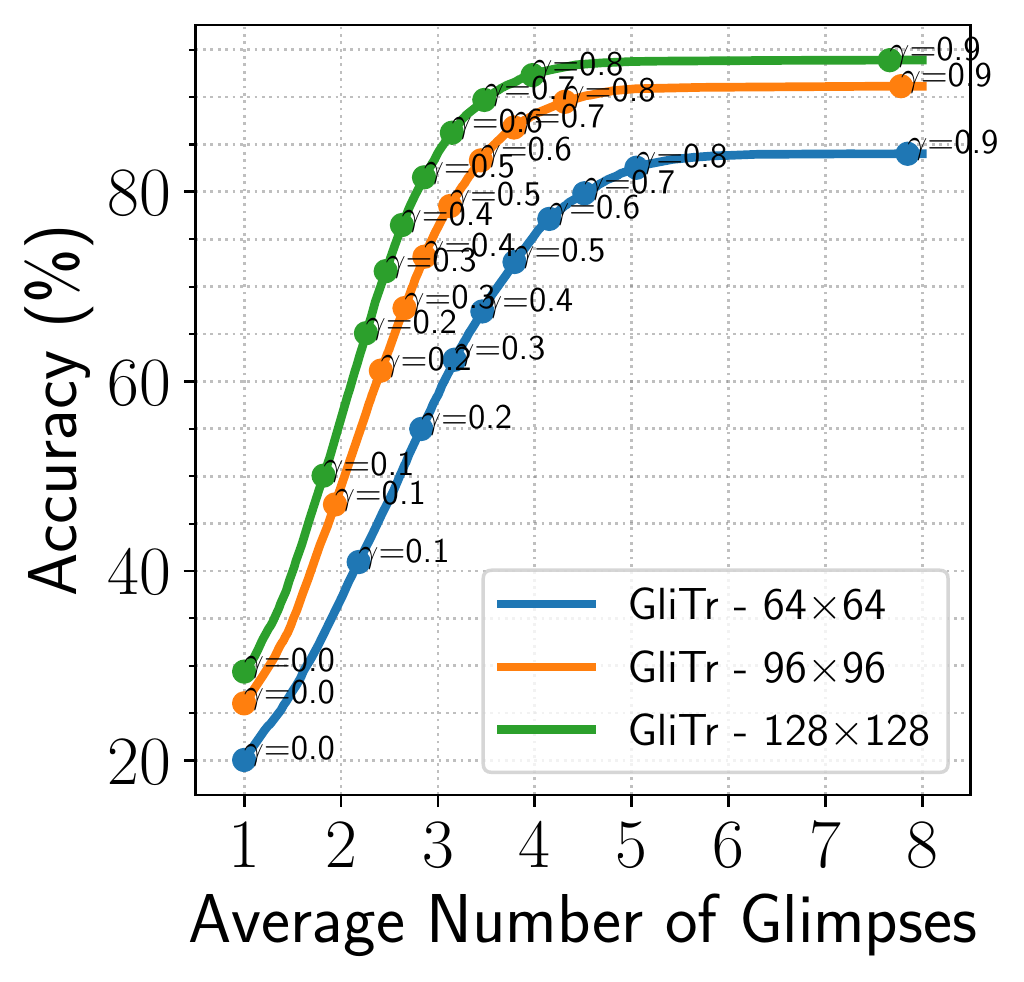}
      \centering{(b)}
    \end{minipage}
    \caption{\textbf{\glitter with early exit.} We display accuracy vs an average number of glimpses seen by \glitter per video to predict a class with probability $>\gamma$. (a) SSv2 and (b) Jester.}
    \label{fig:earlystop}
\end{figure}

We extend \glitter for applications that require timely decision-making. We terminate sensing and conclude a class when \glitter makes a sufficiently confident prediction. We evaluate confidence using the maximum value in the predicted class logits, $\mathcal{C}_t = \max(p(\hat{y}_t))$ and exit when \glitter achieves confidence $\mathcal{C}_t > \gamma$. We show the performance of \glitter for varying $\gamma$ in Figure \ref{fig:earlystop}. We observe a trade-off between the glimpse size and the average number of glimpses required for confident prediction. \glitter achieves higher confidence early with larger glimpse sizes and thus requires fewer glimpses to achieve certain performance. While continued sensing improves \glitter's performance on SSv2, the performance saturates on Jester after the initial $50 \%$ of the glimpses, rendering further sensing unnecessary.

\section{Conclusions}
We develop a novel online action prediction model called Glimpse Transformer (\glitter) that observes video frames only partially through glimpses and predicts an ongoing action solely based on spatially and temporally incomplete observations. It predicts an informative glimpse location for a current frame based on the glimpses observed in the past. Without any ground truth for the glimpse locations, we train \glitter using a novel spatiotemporal consistency objective. On the Something-Something-v2 (SSv2) dataset, the proposed consistency objective yields around $10\%$ higher accuracy than the cross-entropy-based baseline objective. Further, we establish that spatial consistency helps \glitter learn a better glimpse selection strategy, whereas temporal consistency improves classification performance under partial observability. While never observing frames completely, \glitter achieves 53.02\% and 93.91\% accuracy on SSv2 and Jester datasets and reduces the sensing area per frame by $\sim 67\%$. Finally, we also showcase a trade-off between the glimpse size and the number of glimpses required for early action prediction. \glitter is useful for lightweight, low-cost devices with small field-of-view cameras.

{\small
\bibliographystyle{ieee_fullname}
\bibliography{egbib}
}

\newpage
\appendix
\section{Additional Results}
\label{sec:appendix}
\begin{figure}[!h]
    \centering
    \begin{minipage}{0.49\linewidth}
    \includegraphics[page=1,trim={0 0 0 0},clip,width=\linewidth]{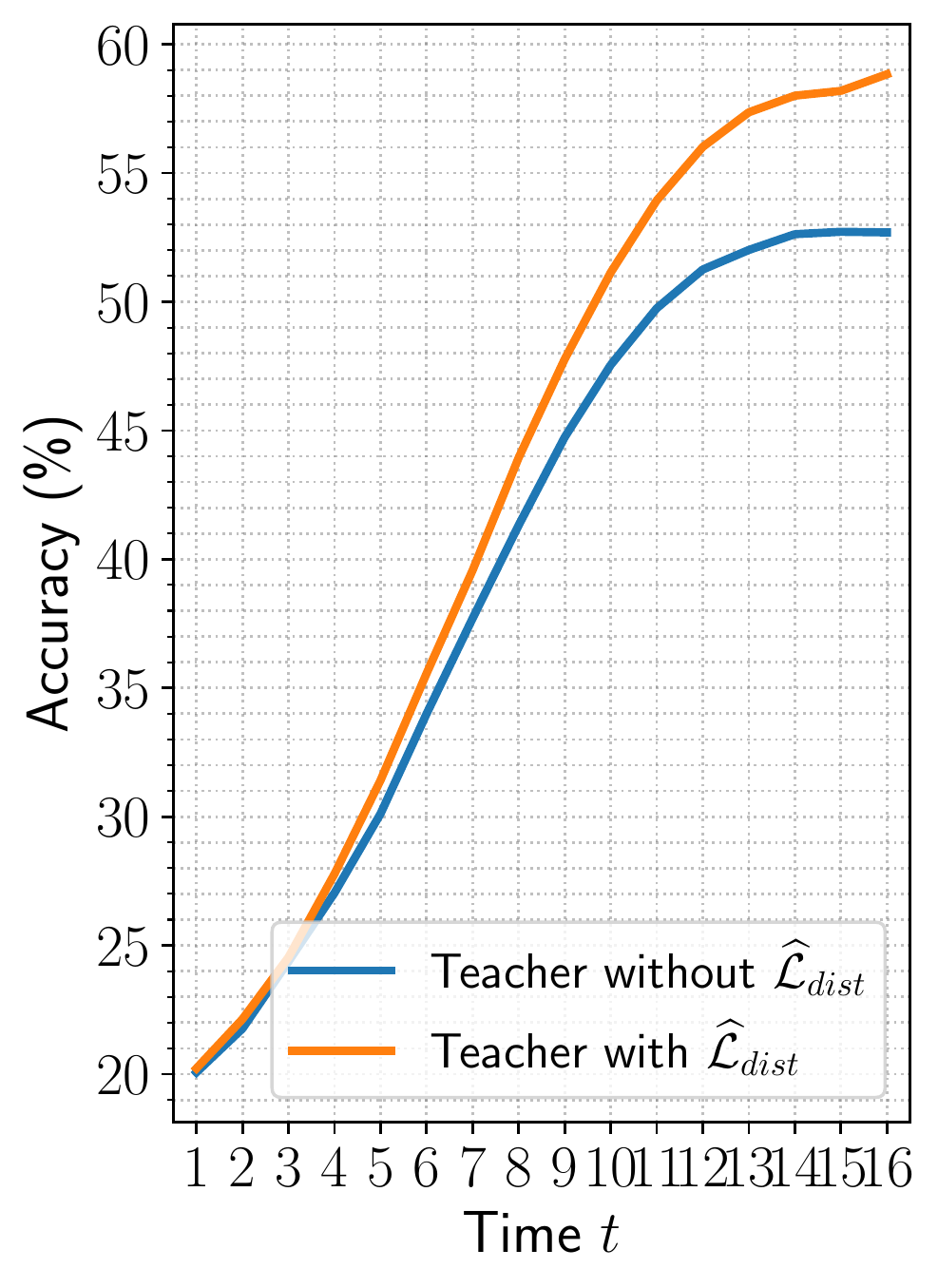}
    \centering{(a)}
    \end{minipage}
    \hfill
    \begin{minipage}{0.49\linewidth}
    \includegraphics[page=1,trim={0 0 0 0},clip,width=\linewidth]{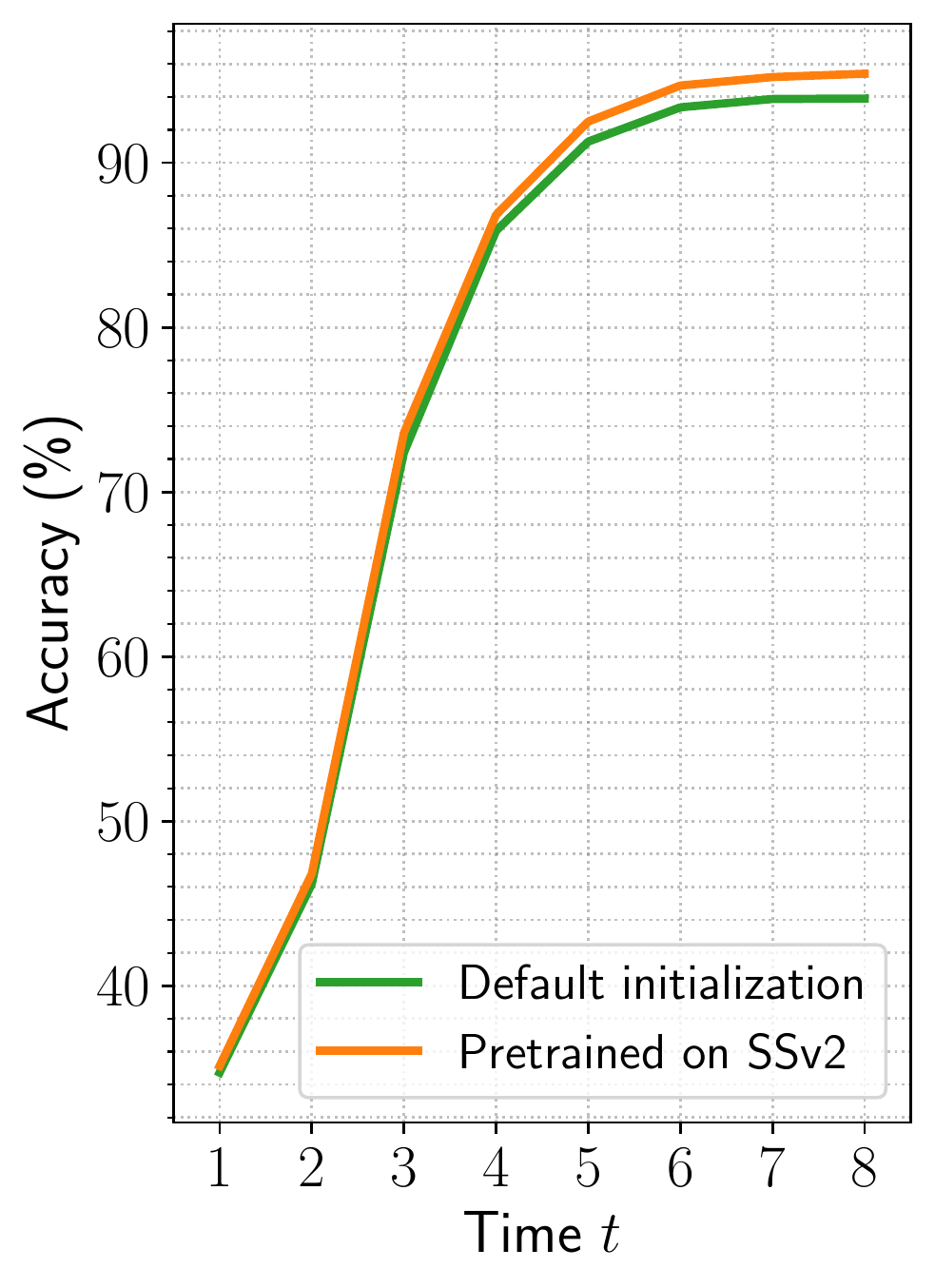}
    \centering{(b)}
    \end{minipage}
    \caption{\textbf{(a) Ablation on $\widetilde{\mathcal{L}}_{dist}$} objective for the teacher trained on SSv2 dataset. (b) \textbf{Ablation on initialization scheme} for the teacher trained on Jester dataset.}
    \label{fig:ablation_teacher}
\end{figure}

\minisection{Ablation on $\widetilde{\mathcal{L}}_{dist}$.}
We distill VideoMAE~\cite{tong2022videomae} (a transformers-based offline action recognition model) to our teacher model on SSv2 dataset. To do so, we minimize $\widetilde{\mathcal{L}}_{dist}$ \ie KL-divergence between the class distributions predicted by our teacher model and VideoMAE based on complete video (equation 6 in main paper). To assess importance of this objective, we train our teacher model with and without $\widetilde{\mathcal{L}}_{dist}$ and display results in Figure \ref{fig:ablation_teacher}(a). We observe improvement of approximately 6\% in accuracy at $t=16$ when $\widetilde{\mathcal{L}}_{dist}$ is included in the training objectives. Note, since a pretrained VideoMAE~\cite{tong2022videomae} in unavailable for Jester, we do not use $\widetilde{\mathcal{L}}_{dist}$ for training the teacher model on this dataset.

\minisection{Ablation on Initialization Scheme.} To improve the performance of the teacher model on the Jester dataset, we initialize its parameters using the parameters of the teacher model pretrained on SSv2 with a complete set of training objectives (equation 9 in the main paper), including $\widetilde{\mathcal{L}}_{dist}$. We compare the performance of the above model with the performance of the teacher initialized using default scheme \ie $\mathcal{T}_f$ initialized using an open-source ViT-S model~\cite{zhou2021image} pretrained on the ImageNet, and $\mathcal{T}_c$ and $\mathcal{T}_l$ initialized randomly. The result shown in Figure \ref{fig:ablation_teacher}(b) indicates that once finetuned on Jester dataset, the teacher pretrained on SSv2 achieves higher performance than the teacher initialized with default scheme, especially for $t>4$. Finally, at $t=8$, the teacher with pretrained weights achieves nearly 1.5\% higher accuracy.

\begin{figure*}
    \centering
    \centering{\small$t=0~~~~~t=1~~~~~t=2~~~~~t=3~~~~~t=4~~~~~t=5~~~~~t=6~~~~~t=7~~~~~t=8~~~~~t=9~~~~t=10~~t=11~~t=12~~t=13~~t=14~~t=15$}
    \begin{minipage}{0.98\linewidth}
    \includegraphics[page=1,trim={0 0 0 0},clip,width=\linewidth]{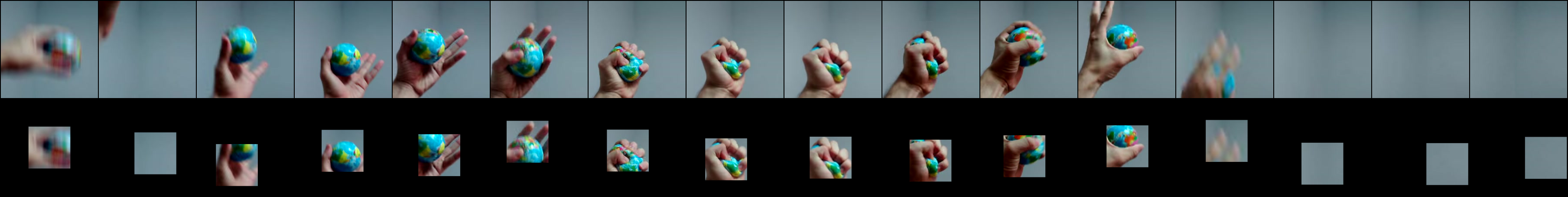}
    \par\vspace{-1.5mm}
    \centering{\small$\hat{y}_{1:9}$=Throwing something in the air and catching it;~~~~ $\hat{y}_{10:16}$=Squeezing something;~~~~ $y_{GT}$=Squeezing something}
    \par\vspace{1mm}
    \end{minipage}
    \vfill
    \begin{minipage}{0.98\linewidth}
    \includegraphics[page=1,trim={0 0 0 0},clip,width=\linewidth]{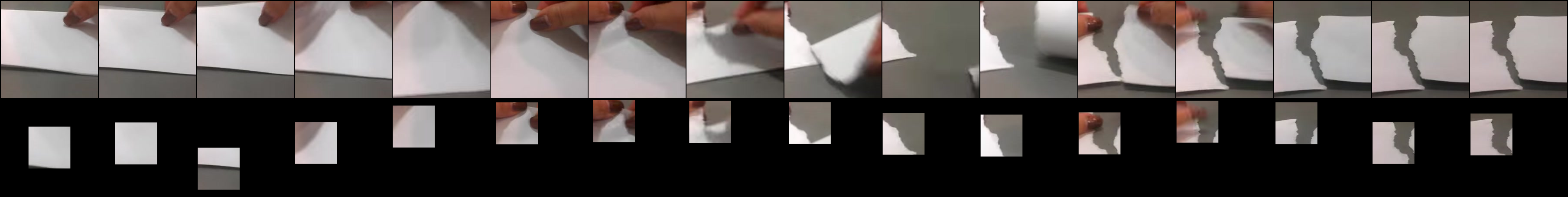}
    \par\vspace{-1.5mm}
    \centering{\small$\hat{y}_{1:4}$=Letting something roll down a slanted surface;~~~~$\hat{y}_{5:16}$=Tearing something into two pieces;~~~~$y_{GT}$=Tearing something into two pieces}
    \par\vspace{1mm}
    \end{minipage}
    \vfill
    \begin{minipage}{0.98\linewidth}
    \includegraphics[page=1,trim={0 0 0 0},clip,width=\linewidth]{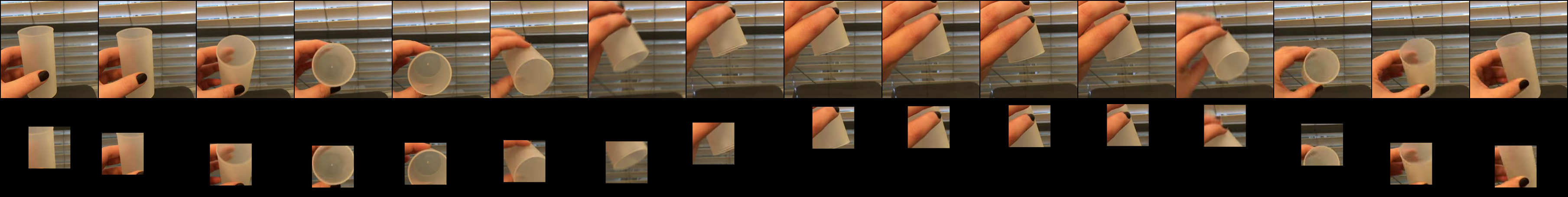}
    \par\vspace{-1.5mm}
    \centering{\small$\hat{y}_{1:2}$=Moving something up;~~~~$\hat{y}_{3:16}$=Showing that something is empty;~~~~$y_{GT}$=Showing that something is empty}
    \par\vspace{1mm}
    \end{minipage}
    \vfill
    \begin{minipage}{0.98\linewidth}
    \includegraphics[page=1,trim={0 0 0 0},clip,width=\linewidth]{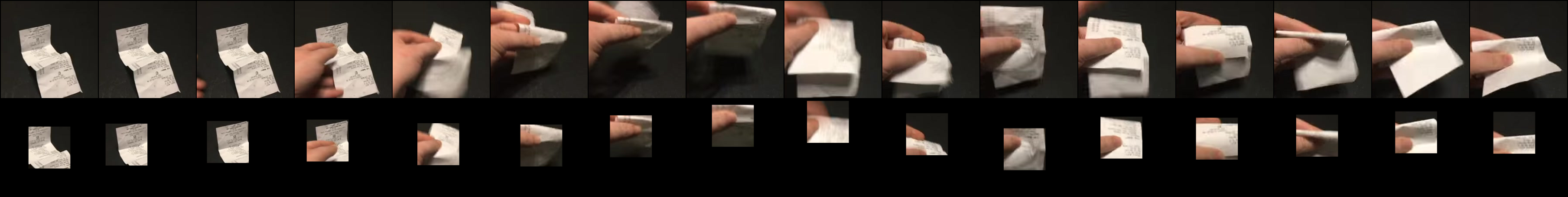}
    \par\vspace{-1.5mm}
    \centering{\small$\hat{y}_{1:4}$=Taking one of many similar things on the table;~~~~$\hat{y}_{5:16}$=Folding something;~~~~$y_{GT}$=Folding something}
    \par\vspace{1mm}
    \end{minipage}
    \vfill
    \begin{minipage}{0.98\linewidth}
    \includegraphics[page=1,trim={0 0 0 0},clip,width=\linewidth]{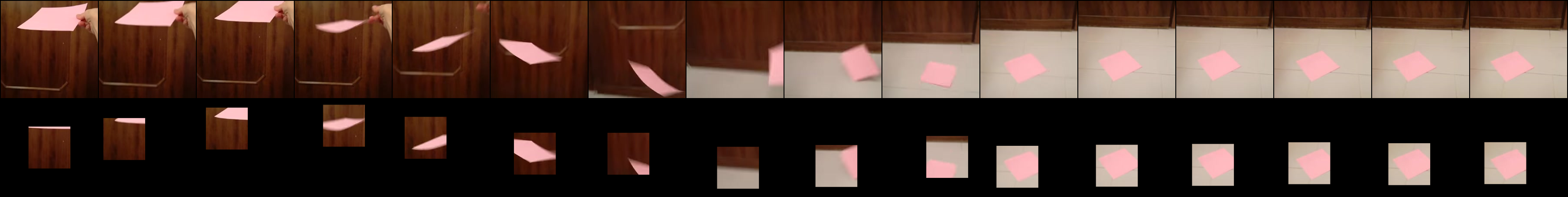}
    \par\vspace{-1.5mm}
    \centering{\small$\hat{y}_{1}$=Moving something down;~~~~$\hat{y}_{2:16}$=Something falling like a feather or paper;~~~~$y_{GT}$=Something falling like a feather or paper}
    \par\vspace{1mm}
    \end{minipage}
    \vfill
    \begin{minipage}{0.98\linewidth}
    \includegraphics[page=1,trim={0 0 0 0},clip,width=\linewidth]{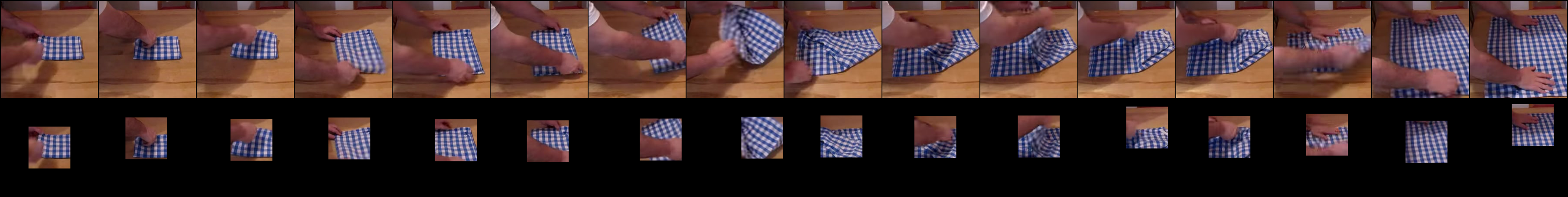}
    \par\vspace{-1.5mm}
    \centering{\small$\hat{y}_{1:8}$=Folding something;~~~~$\hat{y}_{9:16}$=Unfolding something;~~~~$y_{GT}$=Unfolding something}
    \par\vspace{1mm}
    \end{minipage}
    \vfill
    \begin{minipage}{0.98\linewidth}
    \includegraphics[page=1,trim={0 0 0 0},clip,width=\linewidth]{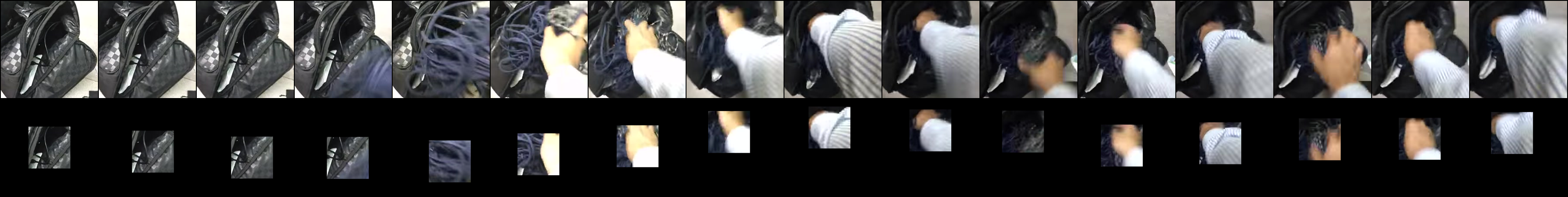}
    \par\vspace{-1.5mm}
    \centering{\small$\hat{y}_{1:3}$=Poking something so lightly that it doesn't or almost doesn't move;~~~~$\hat{y}_{4:16}$=Stuffing something into something;\\$y_{GT}$=Stuffing something into something}
    \par\vspace{1mm}
    \end{minipage}
    \vfill
    \begin{minipage}{0.98\linewidth}
    \includegraphics[page=1,trim={0 0 0 0},clip,width=\linewidth]{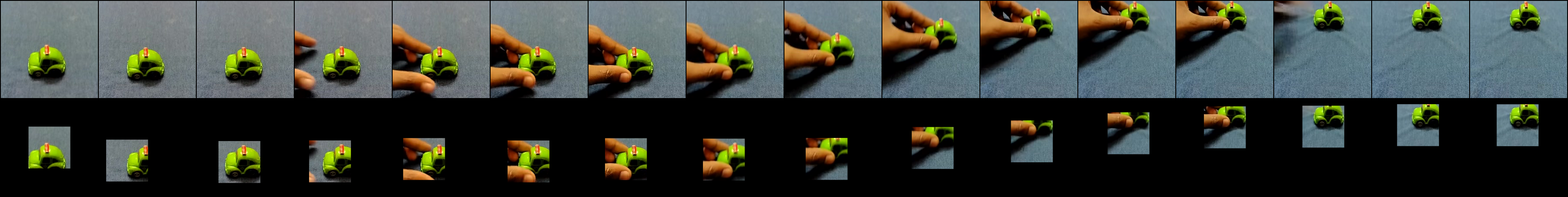}
    \par\vspace{-1.5mm}
    \centering{\small$\hat{y}_{1:3}$=Covering something with something;~~~~$\hat{y}_{4:7}$=Pulling something from right to left;~~~~$\hat{y}_{8:16}$=Moving something up;\\ $y_{GT}$=Moving something up}
    \par\vspace{1mm}
    \end{minipage}
    \vspace{-4mm}
    \caption{Visualization of glimpses (bottom rows) selected by \glitter on SSv2 dataset. Complete frames (top rows) are shown for reference.}
    \label{fig:ssv2_sm_viz}
\end{figure*}

\minisection{Visualization.} We show more visual results on example videos from SSv2 in Figure \ref{fig:ssv2_sm_viz}.

\end{document}